\documentclass{article}

\usepackage[final]{corl_2026}
\usepackage{ulem}
\usepackage{booktabs}
\usepackage{bm}
\usepackage{graphicx}
\usepackage{enumitem}
\usepackage{multirow}
\usepackage{float}
\usepackage{wrapfig}
\usepackage{amssymb}
\usepackage{threeparttable}

\title{FTP-1: A Generalist Foundation Tactile Policy Across Tactile Sensors for Contact-Rich Manipulation}

\author{%
\begin{minipage}{\dimexpr\textwidth-2\tabcolsep\relax}
\centering\normalsize
Chengbo Yuan$^{\ddagger,*,1,2,3}$,
Zicheng Zhang$^{*,3,7}$,
Mingjie Zhou$^{*,3,7}$,\\
Wendi Chen$^{4,8}$,
Yi Wang$^{4,8}$,
Zhuoyang Liu$^{5}$,
Dantong Niu$^{5}$,
Shuo Wang$^{1}$,\\
Hui Zhang$^{6}$,
Wenkang Zhang$^{1}$,
Yingdong Hu$^{1}$,
Yuanqing Gong$^{3}$,
Wanli Xing$^{3}$,\\
Chuan Wen$^{4}$,
Cewu Lu$^{4,8}$,
Kaifeng Zhang$^{\dagger,3}$,
Yang Gao$^{\dagger,1,2}$\\[3pt]
\normalfont\small
$^{1}$Tsinghua University \quad
$^{2}$Shanghai Qi Zhi Institute \quad
$^{3}$Sharpa\\
$^{4}$Shanghai Jiao Tong University \quad
$^{5}$University of California, Berkeley\\
$^{6}$ETH Zurich \quad
$^{7}$Fudan University \quad
$^{8}$Shanghai Innovation Institute\\[3pt]
$^{*}$Equal contribution \quad
$^{\ddagger}$Project Leader \quad
$^{\dagger}$Corresponding author
\end{minipage}
}

\hypersetup{
  pdftitle={FTP-1: A Generalist Foundation Tactile Policy Across Tactile Sensors for Contact-Rich Manipulation},
  pdfauthor={Chengbo Yuan et al.},
  hypertexnames=false
}
\makeatletter
\renewcommand{\@conferencelocation}{Austin, Texas, USA}
\makeatother

\begin{document}
\maketitle

\vspace{-8mm}


\begin{abstract}
Despite the success of vision-based generalist robotic policies, existing tactile-based policies remain tied to fixed embodiments and sensor setups. This is because tactile signals are highly heterogeneous across hardware, making cross-sensor generalization difficult.
We present FTP-1, to the best of our knowledge, \textbf{the first generalist foundation tactile policy pretrained on large-scale heterogeneous tactile manipulation data for finetuning transfer} across diverse sensors and embodiments. 
FTP-1 supports varied tactile inputs, including image-, array-, and state-based signals, by using heterogeneous encoders to project them into unified morphology-aware latent tokens that are jointly modeled by a shared tactile Transformer expert.
Pretrained on $\sim$ 3,000 hours of tactile manipulation data aggregated from 26 data sources, spanning human and robot demonstrations across 21 sensors, FTP-1 learns tactile skills that transfer beyond the sensors seen during pretraining.
Across downstream finetuning experiments spanning 5 hardware configurations, FTP-1 improves contact-rich manipulation on seen sensor setups by +17.2\% and, surprisingly, transfers to two previously unseen tactile-sensor setups, achieving a +31\% gain in success rate.
FTP-1 establishes the first unified foundation baseline for tactile manipulation, providing future tactile policies with a shared model-level starting point. All model checkpoints, datasets, and training code will be released in the future.
\end{abstract}
\begin{center}\small Project page: \url{https://ftp1-policy.github.io/}\end{center}



\section{Introduction}

Generalist robotic policies~\cite{black2024pi_0, bjorck2025gr00t} seek to move robot learning beyond task-specific training~\cite{zhao2023learning, chi2025diffusion} toward large-scale pretrained policy models that can be efficiently adapted to diverse embodiments, tasks, and environments. By scaling data, computation, and model capacity over large-scale heterogeneous datasets, recent vision-based generalist policies, such as $\pi_{0.5}$~\cite{intelligence2025pi_}, have shown that pretraining provides powerful initialization for downstream manipulation tasks.

However, this generalist paradigm remains largely underexplored in tactile-based policy learning. This is a critical gap, as tactile sensing plays an essential role in contact-rich, fine-grained manipulation. Recent pretrained tactile policies~\cite{cheng2025omnivtla, liu2025vtdexmanip, zheng2026omnivta} have begun to scale tactile data for policy pretraining, but remain tied to limited sensor configurations, observation formats, and robot embodiments, rather than a generalist policy. This sensor-specific design is fundamentally constrained by the heterogeneity of tactile signals across hardware, where sensors differ in modality, resolution, morphology, and contact response. 
In this work, we ask whether tactile manipulation~\cite{huang2025tactile, heng2025vitacformer} can be learned under a generalist foundation policy paradigm~\cite{kim2024openvla}: can a single tactile policy absorb heterogeneous tactile experience and transfer to sensors and embodiments beyond those seen during pretraining?

\begin{figure}[t]
\centering
  \includegraphics[width=\linewidth]{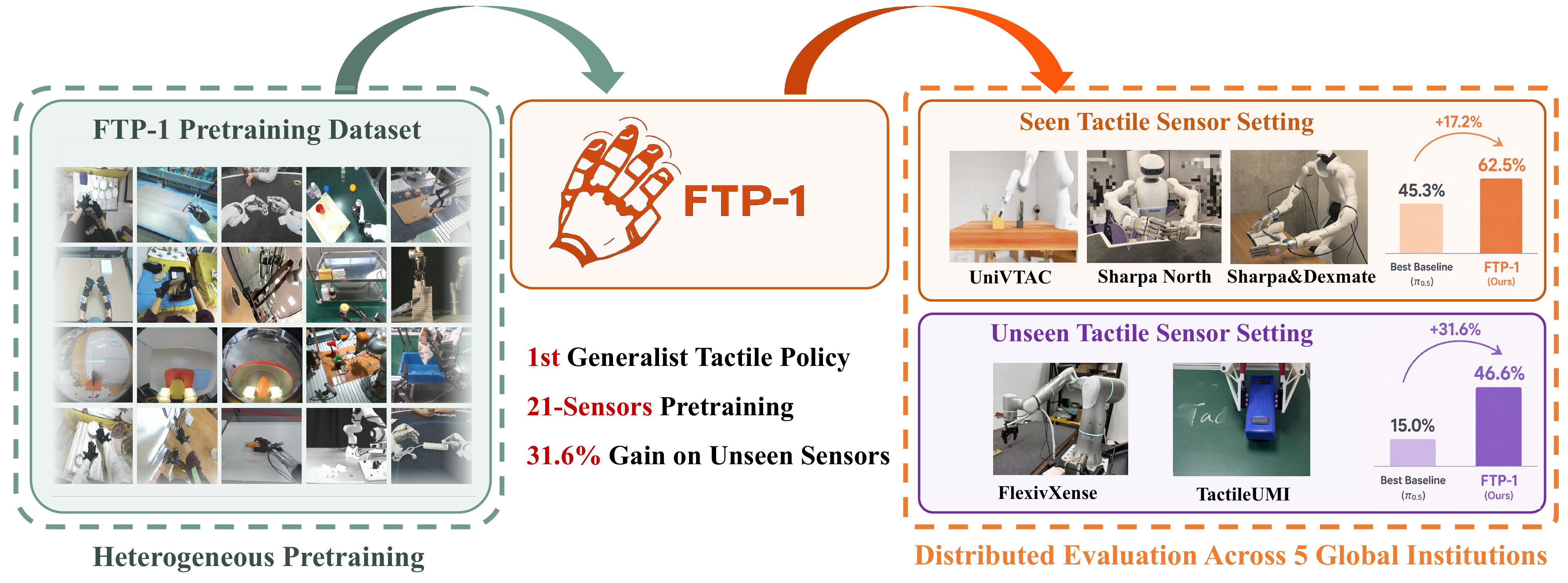}
  \vspace{-3mm}
  \caption{FTP-1 is a generalist foundation tactile policy pretrained for finetuning across diverse sensors and embodiments. Pretrained on a large-scale heterogeneous tactile manipulation dataset, FTP-1 improves downstream performance with a 31.6\% point gain on unseen sensor setups.}
  \label{fig:teaser}
  \vspace{-3.5mm}
\end{figure}

We present \textbf{FTP-1, to the best of our knowledge, the first generalist \underline{F}oundation \underline{T}actile \underline{P}olicy pretrained on large-scale heterogeneous tactile manipulation data for finetuning transfer across diverse sensors and embodiments} (Fig.~\ref{fig:teaser}). Here, a generalist tactile foundation policy is a single action-generating model pretrained across heterogeneous sensors and embodiments, with shared components that can be adapted to new hardware and tasks. FTP-1 supports image-, array-, and state-based tactile inputs. To address tactile heterogeneity, we introduce Morphology-Aware Tactile Token Space (MTTS) as a unified interface across sensors. MTTS represents tactile inputs with a fixed set of functional-area tokens, each augmented by a shared function-area embedding. Heterogeneous tactile encoders are then used to project different tactile inputs into this common space. On top of MTTS, FTP-1 adopts a multi-expert foundation-policy architecture~\cite{black2024pi_0, intelligence2025pi_}, where a shared tactile Transformer expert~\cite{vaswani2017attention} jointly models tactile tokens to learn reusable tactile representations and manipulation skills across sensors.

We further curate FTP-1-Dataset, a new large-scale heterogeneous tactile manipulation dataset aggregated from 26 data sources. It contains approximately 3,000 hours of human and robot demonstrations across 21 tactile sensors, with all data standardized through our unified MTTS tactile interface.
Pretraining on this dataset enables FTP-1 to acquire transferable tactile manipulation skills across sensors and embodiments. 

To evaluate this capability, we distribute FTP-1 pretrained checkpoints to 5 independent institutions spanning global regions, each conducting finetuning on a distinct hardware setup, covering 4 tactile sensors in total. The task suite comprises 14 diverse tasks, covering a broad range of contact-rich behaviors, including in-hand adjustment, force-controlled pressing, insertion and extraction, and long-horizon fine-grained dexterous manipulation.
Results show that FTP-1 improves success rates by +17.2\% on 3 seen sensor setups and, surprisingly, also transfers to previously 2 unseen tactile sensors, achieving a +31\% success-rate gain. Ablation studies further confirm that these gains stem from FTP-1’s transferable tactile manipulation skills. In summary, our contributions are:

\begin{itemize}[leftmargin=35pt, itemsep=2pt, topsep=1pt, parsep=0pt, partopsep=0pt]
    \item \textbf{FTP-1, a generalist foundation tactile policy} pretrained for finetuning transfer across diverse sensors and embodiments, through morphology-aware tactile tokenization and shared tactile expert modeling.
    \item \textbf{A large-scale aggregated tactile manipulation dataset} aggregated from 26 data sources, spanning human and robot data and 21 distinct tactile sensors. By pretraining on this, FTP-1 establishes a shared foundation starting point for tactile manipulation policy training.
    \item \textbf{Results of tactile manipulation knowledge transfer.} Experiments show that FTP-1 improves downstream contact-rich manipulation, even transfer to unseen tactile-sensor setups with 31\% success rate improvement.
\end{itemize}

\vspace{-2mm}
\section{FTP-1: Generalist Foundation Tactile Policy}
\label{sec:method}
\vspace{-3mm}

We introduce FTP-1, a generalist tactile foundation policy for contact-rich manipulation across diverse tactile sensors and robot embodiments. Given a language instruction $\ell$, (multi-view) RGB observations $\mathcal{I}_t$, proprioception $\mathbf{s}_t$, and tactile observations $\mathcal{X}_t$, FTP-1 predicts an action chunk
$\hat{\mathbf{A}}_{t:t+H-1}=\pi_\theta(\ell,\mathcal{I}_t,\mathbf{s}_t,\mathcal{X}_t)$,
where $\hat{\mathbf{A}}_{t:t+H-1}\in\mathbb{R}^{H\times D}$, $H$ is the action horizon, and $D$ is the dimension of a predefined Unified Action Space (UAS)~\cite{zhang2026unidex} that handles heterogeneity of control signal, which is detailed in App.~\ref{app:UAS}.
We next describe FTP-1's key components: the Morphology-Aware Tactile Token Space (MTTS) definition as the unified tactile interface (Sec.~\ref{sec:method_mtts}), heterogeneous tactile encoders for MTTS tokenization (Sec.~\ref{sec:method_tactile_encoder}), a shared tactile expert for modality fusion (Sec.~\ref{sec:method_architecture}), and the pretraining dataset (Sec.~\ref{sec:method_dataset}).

\begin{figure}[t]
\centering
  \includegraphics[width=\linewidth]{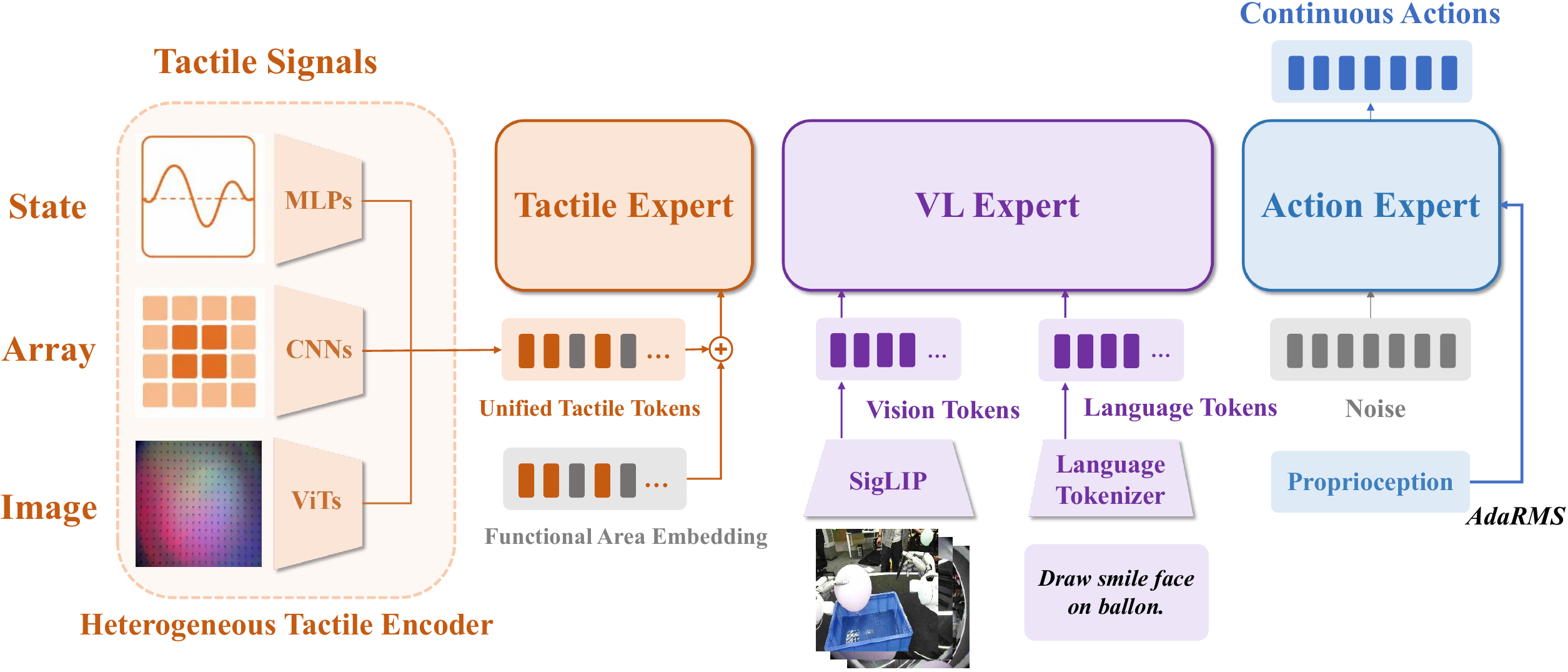}
  \vspace{-3mm}
  \caption{Overview of FTP-1 architecture. Heterogeneous tactile observations are mapped into a unified tactile token space through sensor-specific encoders, processed by a shared tactile expert, and fused with vision-language and proprioceptive information for action generation.}
  \label{fig:method_architecture}
  \vspace{-1mm}
\end{figure}

\vspace{-1mm}

\subsection{Morphology-Aware Tactile Token Space (MTTS)}
\label{sec:method_mtts}

\vspace{-1mm}

\begin{wrapfigure}[12]{r}{0.3\linewidth}
    \centering
    \vspace{-0.6em}
    \includegraphics[width=0.85\linewidth]{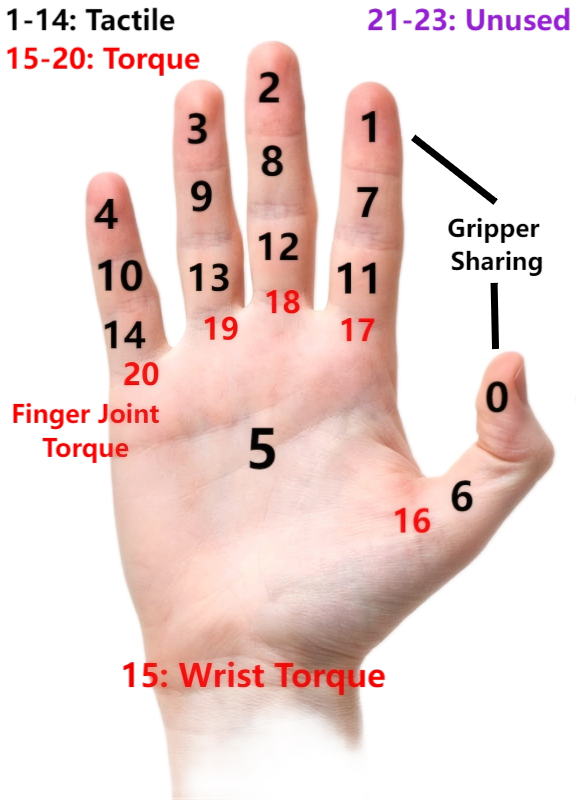}
    \vspace{-3mm}
    \caption{\small Tactile functional-area definition of MTTS.}
    \label{fig:mtts_human_hand}
    \vspace{-1.0em}
\end{wrapfigure}

Tactile sensors vary in modality, resolution, and morphology, making cross-sensor representation sharing challenging. FTP-1 addresses this challenge with a Morphology-Aware Tactile Token Space (MTTS), augmented by shared functional-area embeddings. MTTS organizes tactile signals into 24 functional areas~\cite{niu2026learning}, each represented by one token; the full definition is shown in Fig.~\ref{fig:mtts_human_hand}. For in-hand tactile signals, we use slots 0-14 to represent different hand functional regions. For force/torque signals from wrists and fingers, we use slots 15-20. For parallel grippers, the two gripper-side sensors are mapped to the thumb-tip slot (slot 0) and index-fingertip slot (slot 1), reflecting their two-finger grasping function. Slots 21-23 are reserved for future use.

Under this definition, sensor-specific tactile signals are grouped by functional area and encoded into unified morphology-aware tactile tokens. To indicate the functional area of each token, we add a learnable functional-area embedding, shared across all sensors, before feeding the tokens into the subsequent architecture. We use separate functional-area embeddings for the left/right hands to distinguish their tactile tokens.

\subsection{Heterogeneous Tactile Encoders for MTTS Tokenization}\label{sec:method_tactile_encoder}

Although MTTS provides a unified token interface, tactile inputs from different sensors still vary in shape and modality. We therefore use heterogeneous tactile encoders~\cite{wang2024scaling} to tokenize sensor-specific inputs into MTTS. For each sensor, we first group its signals according to MTTS functional areas, then categorize each group into one of three observation types: image, array, or state.

For image-type inputs (e.g., GelSight~\cite{yuan2017gelsight}), we use a lightweight sensor-specific ViT~\cite{dosovitskiy2020image} followed by a shared pretrained T3 Transformer tactile encoder~\cite{zhao2024transferable} across sensors, and use the final \texttt{[CLS]} token as the tactile token. For array-type inputs (e.g., Contactile~\cite{velasco2025touch}), we use a CNN~\cite{wu2017introduction} to capture spatial tactile structure and compress each functional area into one token. For state-type inputs (e.g., force-torque), we Fourier-encode the raw state~\cite{huang2025spatially} and process it with a lightweight MLP. For sensors with multiple functional areas of the same observation shape, we share the corresponding encoder across these areas to reduce sensor-specific parameters and encourage common tactile dynamics to be modeled consistently. Details are provided in App.~\ref{app:tactile_encoder}.

\begin{figure}[t]
\centering
  \includegraphics[width=\linewidth]{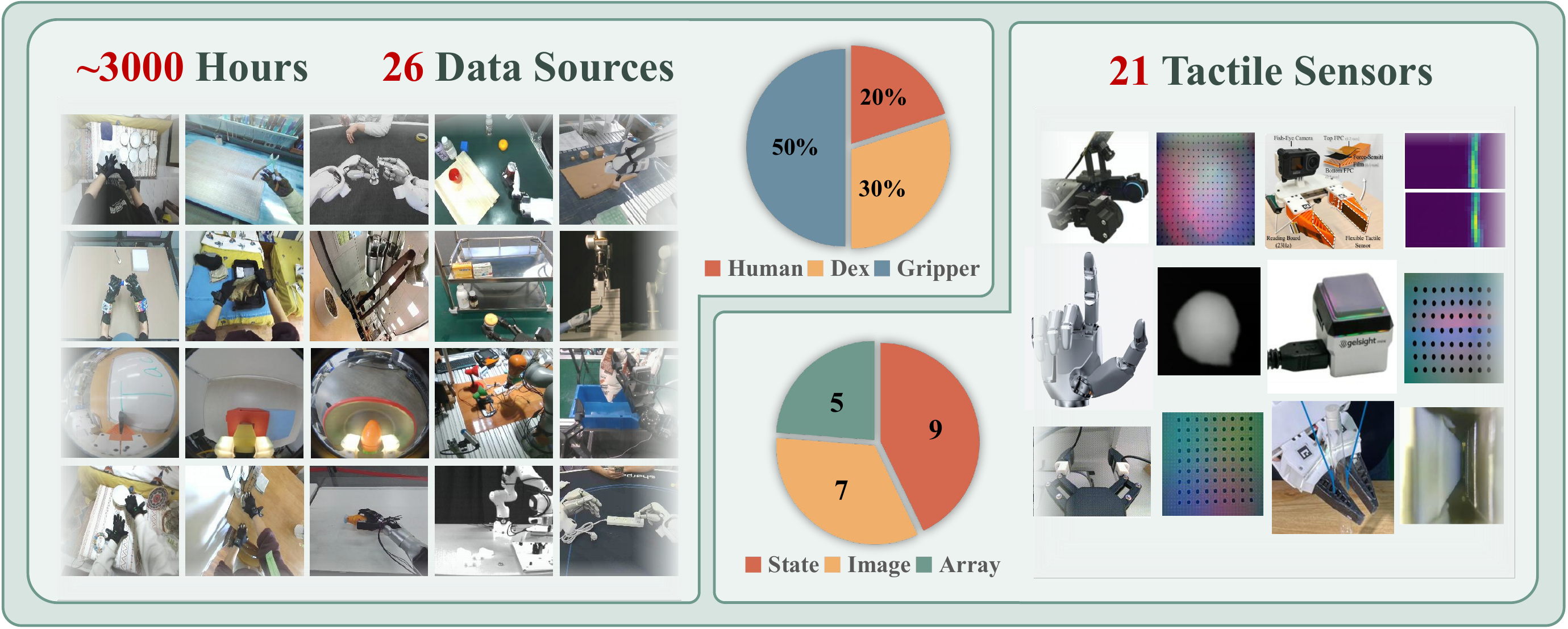}
  \caption{Overview of the FTP-1-Dataset. The dataset aggregates 26 sources across human and robot manipulation, covering 21 tactile sensors with image-, array-, and state-based modalities, all organized under the MTTS interface for unified pretraining.}
  \label{fig:datasets_main}
  \vspace{-5mm}
\end{figure}

\subsection{Shared Tactile Expert for Modality Fusion}
\label{sec:method_architecture}

Finally, FTP-1 adopts a multi-expert architecture~\cite{black2024pi_0} to fuse tactile tokens with vision-language perception for action generation, as shown in Fig.~\ref{fig:method_architecture}. Built on $\pi_{0.5}$~\cite{intelligence2025pi_}, FTP-1 processes image observations and language instructions with a pretrained vision-language Transformer expert~\cite{beyer2024paligemma}, whose outputs are attended by a flow-matching Transformer action expert~\cite{black2024pi_0}. We further fuse proprioceptive state through adaptive RMSNorm~\cite{intelligence2025pi_}: an MLP encodes the Fourier features of proprioception, which are normalized and concatenated with flow-matching timestep features to condition the attention blocks. Details are provided in App.~\ref{app:proprioception_injector}.

Unlike prior tactile-augmented VLA models that inject tactile inputs into the vision-language expert via lightweight adapters~\cite{zhang2023llama, cheng2025omnivtla, huang2025tactile, li2026forcevla2}, FTP-1 uses an independent tactile expert for extracted tactile tokens. This design (1) supports reuse pretrained shared tactile expert on unseen sensors during finetuning (Section~\ref{sec:transfer-to-unseen}), promotes transferable tactile manipulation skills, and (2) avoids disturbing pretrained vision-language knowledge. It also (3) improves tactile processing efficiency and contact-rich manipulation performance, as shown by our comparison with adapter-based methods such as Tactile-VLA~\cite{huang2025tactile} (Sec.~\ref{sec:main-experiment} and Sec.~\ref{sec:transfer-to-unseen}). For fusion, queries from the action expert attend to keys and values from the tactile expert together with the vision-language context. The vision-language and tactile experts each attend within their own token stream; action tokens attend to all three streams. This block attention mask lets tactile information condition action generation directly while preserving separate vision-language and tactile processing. The connectivity is detailed in App.~\ref{app:attention_connectivity}. The tactile expert is a 300M-parameter Transformer~\cite{vaswani2017attention}. 
More complex modality fusion designs e.g. MoE-based fusion~\cite{fang2026force, li2026forcevla2, tang2026towards}, did not yield consistent gains in our experiment, so we adopt the simplest multi-expert design.
Together, these choices provide a simple modular architecture for learning transferable tactile manipulation skills across sensors and embodiments.


\subsection{FTP-1 Pretraining Datasets}\label{sec:method_dataset}

FTP-1 is pretrained on a large-scale heterogeneous tactile manipulation dataset (Fig.~\ref{fig:datasets_main}) aggregated from 26 sources, covering 21 distinct tactile sensors: 7 image-type, 5 array-type, and 9 state-type sensors. In addition to aggregating existing datasets, we collect Sharpa North-FTP-1 with 4,000 long-horizon dexterous demonstrations. All tactile annotations are organized according to MTTS functional areas, described in Sec.~\ref{sec:method_mtts}. To mitigate data imbalance, we apply dataset-specific sampling ratios during training; after re-sampling, the proportions of human, dexterous-hand robot, and gripper robot data are approximately 20\%, 30\%, and 50\%, respectively. The final aggregated dataset contains around 3,000 hours of tactile manipulation data. Details are provided in App.~\ref{app:pretrain_dataset_and_sampling}.


\begin{figure}[t]
\centering
  \includegraphics[width=\linewidth]{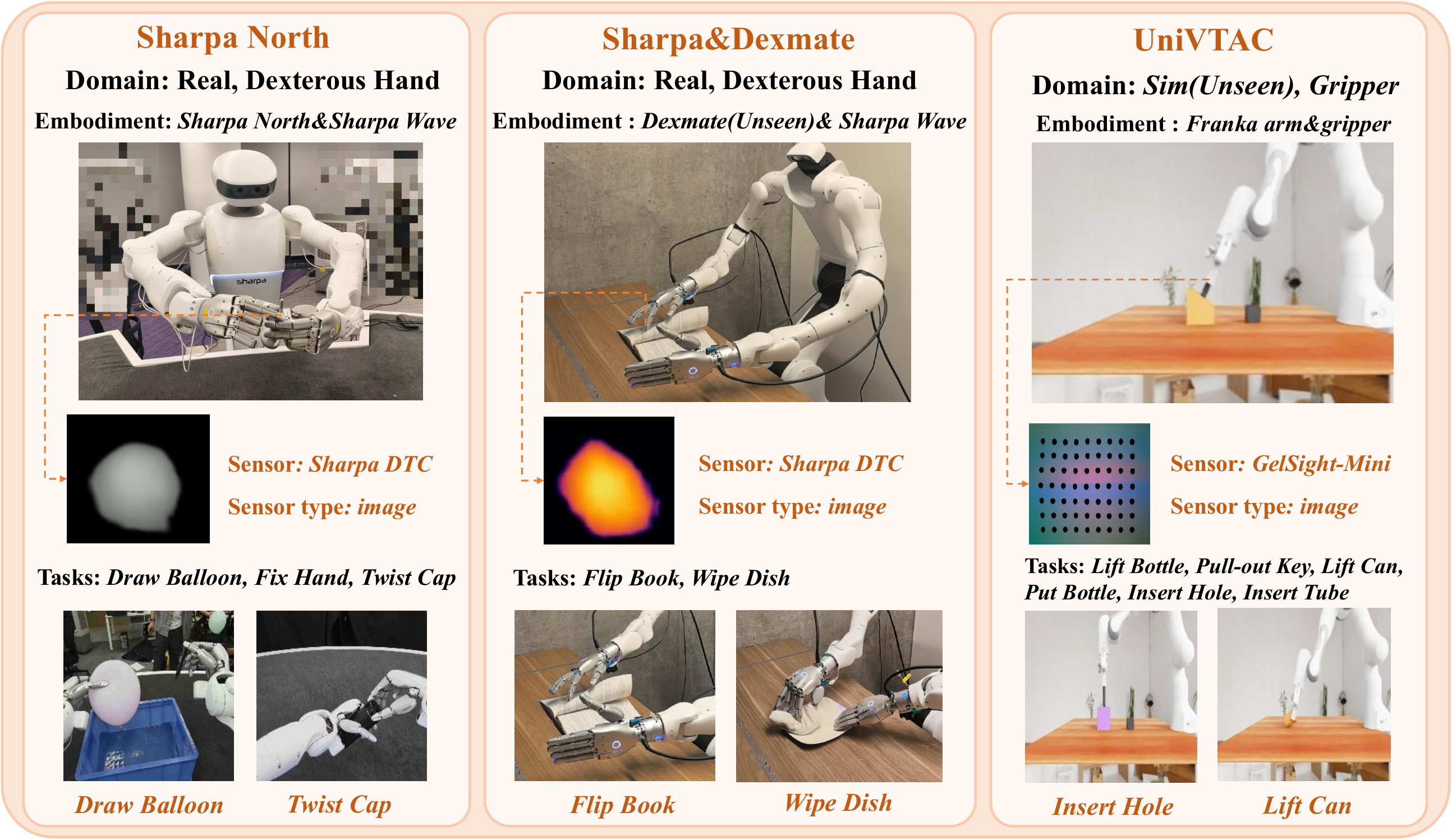}
  \vspace{-3mm}
  \caption{Overview of the finetuning experiment for seen sensors during pretraining.}
  \label{fig:eval_setting_seen}
  \vspace{-5mm}
\end{figure}

\vspace{-3mm}
\section{Finetuning Experiment}\label{sec:main-experiment}
\vspace{-3mm}

In this section, we address the central question: \textbf{Does FTP-1 pretraining improve downstream tactile policy finetuning by learning effective representations for tactile sensor perception?} To answer this, we distribute FTP-1 pretrained checkpoint to 3 independent institutions, each evaluates on a distinct embodiment and task suite across simulation and real-robot settings. Training details are provided in App.~\ref{app:training_details}. We next describe the evaluation setup and results.

\vspace{-3mm}
\subsection{Baselines and Metrics}
\vspace{-3mm}

For simulation experiments, we evaluate 100 rollouts per task, following the original benchmark~\cite{chen2026univtac}. For real-robot experiments, we evaluate 20 rollouts per task. We report finetuned success rates and compare the pretrained FTP-1 with three baselines:

\begin{itemize}[leftmargin=35pt, itemsep=2pt, topsep=0pt, parsep=0pt, partopsep=0pt]
    \item \textbf{\boldmath$\pi_{0.5}$}~\cite{intelligence2025pi_}: a SOTA open-source VLA model without tactile input. This baseline evaluates the benefit of tactile sensing over a strong vision-language-action policy.

    \item \textbf{Tactile-VLA}~\cite{huang2025tactile}: a tactile-based VLA architecture adapted from \citet{huang2025tactile}, which injects tokenized tactile inputs into the VLM expert without a separate tactile expert. This baseline evaluates the effect of our tactile-expert design.

    \item \textbf{FTP-{\boldmath$\pi_{0.5}$}}: our architecture partially initialized with $\pi_{0.5}$ weights but without FTP-1 pretraining. This baseline isolates the contribution of large-scale tactile pretraining.
\end{itemize}

These baselines are chosen to disentangle the effects of tactile input, tactile integration architecture, and FTP-1 pretraining. All methods are implemented based on the $\pi_{0.5}$~\cite{intelligence2025pi_} codebase with comparable model scale, finetuning data, and training protocol for for a fair comparison. 

\subsection{Task Settings}

We evaluate FTP-1 on three embodiments: UniVTAC~\cite{chen2026univtac} in simulation, and Sharpa North and Sharpa\&Dexmate in the real world. An overview of the evaluation setups is shown in Fig.~\ref{fig:eval_setting_seen}. These setups cover two tactile sensors included in pretraining, GelSight-Mini~\cite{yuan2017gelsight} and Sharpa DTC~\cite{heng2025vitacformer}, so both the tactile tokenizer and tactile expert of our methods are initialized from FTP-1 checkpoints. For UniVTAC, we choose six contact-rich tasks spanning in-hand manipulation and contact-aware insertion/extraction; Sharpa North focuses on three long-horizon dexterous manipulation tasks: Draw Balloon, Fix Hand and Twist Cap, including deformable-object interaction, small-part assembly, and bimanual twisting behaviours. Sharpa\&Dexmate focuses on pressing and force-control behaviors, including Flip Book and Wipe Dish tasks. Hardware setups, task definitions, experiment details, and excluded UniVTAC tasks are provided in App.~\ref{app:hardware_setup} and App.~\ref{app:evaluation_setup_seen_sensor}. Examples of policy rollouts are provided in App.~\ref{app:seen_sensor_rollout}.

\begin{table}[t]
\centering
\caption{Results on the UniVTAC simulation benchmark. We report success rates (\%). ``Avg. w/o Lifts'' excludes Lift Bottle and Lift Can. The best and second-best results are highlighted in \textbf{bold} and \underline{underline}, respectively. Results marked with * are from the original benchmark~\cite{chen2026univtac}.}
\vspace{-1mm}
\label{tab:univtac_results}
\setlength{\tabcolsep}{4pt}
\renewcommand{\arraystretch}{1.08}
\resizebox{\linewidth}{!}{
\begin{tabular}{lcccccccc}
\toprule
\textbf{Method}
& \textbf{Lift Bottle}
& \textbf{Pull-out Key}
& \textbf{Lift Can}
& \textbf{Put Bottle}
& \textbf{Insert Hole}
& \textbf{Insert Tube}
& \textbf{Avg.}
& \textbf{Avg. w/o Lift}\\
\midrule
\textbf{VITaL*}~\cite{george2025vital}
& 72 & \underline{47} & 8 & 32 & 25 & 34 & 36.33 & 34.5 \\
\textbf{UniVTAC-ACT*}~\cite{chen2026univtac}
& 71 & 46 & 29 & 31 & 25 & 56 & 43.00 & 39.5 \\

\textbf{\boldmath$\pi_{0.5}$}~\cite{intelligence2025pi_}
& \textbf{97} & 38 & \textbf{72} & 16 & 31 & 41 & \underline{49.16} & 31.5 \\

\textbf{Tactile-VLA}~\cite{huang2025tactile}
& \textbf{97} & 32 & 15 & 10 & 41 & 56 & 41.83 & 34.75 \\

\textbf{FTP-\boldmath$\pi_{0.5}$}
& \underline{77} & 30 & 26 & \underline{19} & \underline{47} & \underline{72} & 45.16 & \underline{42} \\

\midrule
\textbf{FTP-1}
& \textbf{97} & \textbf{48} & \underline{65} & \textbf{47} & \textbf{64} & \textbf{79} & \textbf{66.66} & \textbf{59.5} \\
\bottomrule
\end{tabular}
}
\end{table}

\begin{table}[t]
\centering
\caption{Real-robot results on seen tactile sensor setups. We report finetuned success rates (\%). The best and second-best results are highlighted in \textbf{bold} and \underline{underline}, respectively.}
\label{tab:real_seen_results}
\vspace{-1mm}
\scriptsize
\setlength{\tabcolsep}{2.5pt}
\renewcommand{\arraystretch}{1.05}
\resizebox{\linewidth}{!}{
\begin{tabular}{lccccccc}
\toprule
\textbf{Method}
& \multicolumn{4}{c}{\textbf{Sharpa North}}
& \multicolumn{2}{c}{\textbf{Sharpa\&Dexmate}}
& \textbf{Average} \\
\cmidrule(lr){2-5} \cmidrule(lr){6-7}
& \textbf{Draw Balloon}
& \textbf{Fix Hand (Tear)}
& \textbf{Fix Hand (Finish)}
& \textbf{Twist Cap}
& \textbf{Flip Book}
& \textbf{Wipe Dish}
& \\
\midrule
\textbf{\boldmath$\pi_{0.5}$}~\cite{intelligence2025pi_}
& \underline{35} & \underline{70} & \underline{35} & \underline{40} & 65 & 30 & \underline{45.3} \\

\textbf{Tactile-VLA}~\cite{huang2025tactile}
& 20 & \textbf{80} & 25 & 10 & 45 & 35 & 35.8 \\

\textbf{FTP-\boldmath$\pi_{0.5}$}
& 25 & 65 & 25 & 20 & \underline{70} & \underline{45} & 41.6 \\

\midrule
\textbf{FTP-1}
& \textbf{45} & \textbf{80} & \textbf{40} & \textbf{65} & \textbf{85} & \textbf{60} & \textbf{62.5} \\
\bottomrule
\end{tabular}
}
\vspace{-3mm}
\end{table}

\subsection{Experiment Results}

\textbf{Simulation Results.} Tab.~\ref{tab:univtac_results} reports the UniVTAC benchmark results~\cite{chen2026univtac}. Surprisingly, Lift Bottle and Lift Can can be largely solved without tactile input in simulation: $\pi_{0.5}$ already achieves 97\% and 72\% success rates, respectively. We therefore additionally report the average success rate excluding these two tasks. FTP-1 achieves the best performance under both metrics, with 66.66\% overall success and 59.5\% excluding the lift tasks, outperforming the second-best method by about +17.5\% in both cases. 

\textbf{Real-World Results.} Results are shown in Tab.~\ref{tab:real_seen_results}. FTP-1 achieves the best performance among all methods, with an average success rate of 62.5\%. Surprisingly, $\pi_{0.5}$ ranks second with a 45.3\% average success rate, outperforming the other two tactile-based baselines. This suggests that, without proper modality fusion, tactile inputs may hurt performance by interfering with the vision-language perception module, consistent with observations in previous work~\cite{niu2026learning}, especially on long-horizon tasks in the Sharpa North domain.

In terms of detailed behavior, Tactile-VLA~\cite{huang2025tactile} and FTP-$\pi_{0.5}$ tends to generate unstable actions when contact conditions change, indicating limited robustness to tactile perception without our tactile-expert fusion design. Without tactile inputs, $\pi_{0.5}$ fails to maintain consistent pressing force on the two Sharpa\&Dexmate tasks sometimes and pushes against the bottle cap without reactive force adjustment. In contrast, FTP-1 produces more stable and smoother actions across these tasks.

\begin{figure}[t]
\centering
  \includegraphics[width=1.0\linewidth]{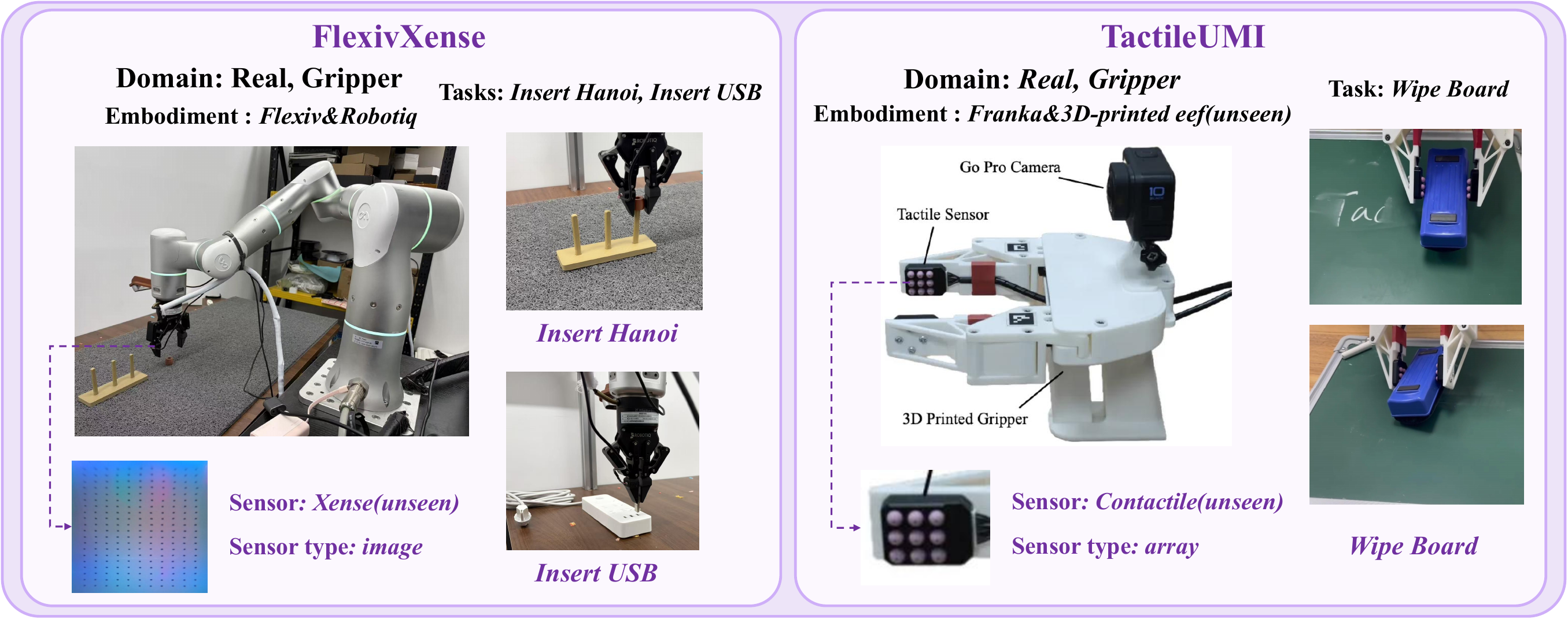}
  \vspace{-5.5mm}
  \caption{Overview of the finetuning experiment for unseen sensors during pretraining.}
  \vspace{-5.5mm}
  \label{fig:unseen_eval_setting}
\end{figure}

\vspace{-3mm}
\section{Experiment of Transferable Tactile Manipulation Abilities}
\vspace{-3mm}

While prior work shows that in-domain tactile-sensor pretraining improves downstream policy performance~\cite{zhao2024transferable, cheng2025omnivtla, zheng2026omnivta}, whether large-scale heterogeneous pretraining enables transfer to unseen sensors remains unclear. We study finetuning transfer of an action-generating policy pretrained on heterogeneous tactile manipulation data. 

Here, an \textbf{unseen sensor} is specific tactile hardware absent from FTP-1 pretraining. Xense and Contactile are new hardware within the image- and array-based modality families covered by pretraining. During downstream finetuning, FTP-1 trains a newly initialized sensor-specific encoder on demonstrations collected with the target sensor, while reusing the pretrained tactile expert and functional-area embeddings, as well as the shared T3 Transformer trunk for image-type inputs~\cite{zhao2024transferable}. The vision-language and action experts are also initialized from the FTP-1 checkpoint. All methods use the same target-task demonstration sets and the finetuning protocol in App.~\ref{app:training_details}. We first evaluate downstream finetuning performance, then conduct ablations to verify whether the gains come from pretrained tactile knowledge.

\vspace{-3mm}
\subsection{Transfer to Unseen Tactile Sensor Setups}\label{sec:transfer-to-unseen}

To evaluate FTP-1 on unseen tactile sensors, we distribute FTP-1 checkpoints to two new institutions with two new setups: FlexivXense with Xense image tactile sensors and TactileUMI~\cite{huang2025tactile} with Contactile array sensors~\cite{velasco2025touch}, shown in Fig.~\ref{fig:unseen_eval_setting}. We also provide the pretraining sensors most similar to these two evaluation sensors in App.~\ref{app:similar_sensor}.
These setups include two fine-grained insertion tasks, Insert Hanoi (inserting a small circle hanoi toy to the pillar) and Insert USB  for the FlexivXense domain (100 finetuning demonstrations per-task), and one press-control task, Wipe Board for the TactileUMI domain (50 demonstrations per-task). Hardware details are provided in App.~\ref{app:hardware_setup}. Examples of policy rollouts are provided in App.~\ref{app:unseen_sensor_rollout}.

\begin{wraptable}{r}{0.48\linewidth}
\vspace{-4mm}
\centering
\caption{Real-robot results on unseen tactile sensor setups with success rate as the metric.}
\label{tab:real_unseen_results}
\vspace{1mm}
\scriptsize
\setlength{\tabcolsep}{1.8pt}
\renewcommand{\arraystretch}{0.95}
\begin{tabular}{lcccc}
\toprule
\textbf{Method}
& \multicolumn{2}{c}{\textbf{FlexivXense}}
& \textbf{TactileUMI}
& \textbf{Avg.} \\
\cmidrule(lr){2-3} \cmidrule(lr){4-4}
& \textbf{Insert Hanoi}
& \textbf{Insert USB}
& \textbf{Wipe Board}
& \\
\midrule
\textbf{\boldmath$\pi_{0.5}$~\cite{intelligence2025pi_}}
& \underline{25} & 0 & 20 & \underline{15.0} \\
\textbf{Tactile-VLA~\cite{huang2025tactile}}
& 0 & \underline{10} & 15 & 8.3 \\
\textbf{FTP-\boldmath$\pi_{0.5}$}
& 5 & \underline{10} & \underline{30} & \underline{15.0} \\
\midrule
\textbf{FTP-1}
& \textbf{55} & \textbf{30} & \textbf{55} & \textbf{46.6} \\
\bottomrule
\end{tabular}
\vspace{-4mm}
\end{wraptable}

\textbf{Evaluation Results.} Tab.~\ref{tab:real_unseen_results} reports the evaluation results on two unseen tactile sensor setups. Our method achieves the highest success rate among all baselines, reaching 46.6\%. Compared with FTP-$\pi_{0.5}$ (15\%), it improves the success rate by +31.6\%, demonstrating that pretraining on large-scale heterogeneous tactile manipulation data effectively improves finetuning performance under unseen sensor settings. In contrast, $\pi_{0.5}$ performs comparably to FTP-$\pi_{0.5}$, suggesting that simply adding a tactile branch does not necessarily help without an appropriate modality-fusion prior and pretraining knowledge.

For Insert Hanoi, FTP-1 and $\pi_{0.5}$ exhibit recovery behaviors, while the other baselines do not. This recovery ability likely benefits from large-scale pretraining. Moreover, FTP-1 shows reactive insertion control: when the circular Hanoi piece is misaligned with the pillar, FTP-1 slows down the insertion motion based on tactile feedback, whereas $\pi_{0.5}$ does not, often leading to failure. For Insert USB, the task is difficult to master with only 100 demonstrations, making it a good test of data efficiency. Compared with FTP-1, other models are less stable and sometimes exhibit small shaking motions during insertion, which reduces their success rates. For Wipe Board, we observe a similar trend to the previous Sharpa\&Dexmate tasks: other models struggle to maintain stable pressing force and may lose tight contact with the board during wiping. Overall, these results show that FTP-1’s large-scale heterogeneous tactile pretraining provides a strong finetuning starting points, enabling better generalization, data efficiency, and reactive control in unseen tactile sensor settings.
    
\vspace{-3mm}
\subsection{Improvements Come from Pretrained Tactile Knowledge}

The previous section shows that FTP-1 provides a strong initialization for tactile policy learning and improves downstream finetuning performance, even for unseen tactile sensors. However, it remains unclear whether these gains truly come from pretrained tactile knowledge. We therefore examine two possible explanations:

\begin{itemize}[leftmargin=35pt, itemsep=2pt, topsep=0pt, parsep=0pt, partopsep=0pt]
\item \textbf{Hypothesis 1 (Data Distribution):} The gains arise because the FTP-1 pretraining data is closer to the downstream task distribution, making finetuning easier.
\item \textbf{Hypothesis 2 (Transferable Knowledge):} FTP-1 learns transferable tactile manipulation knowledge, providing better initialization for downstream tactile policy finetuning.
\end{itemize}

To verify this, we pretrain a No-Tactile-Pretraining checkpoint (NTP) using the same setup as FTP-1, including the same pretraining data and optimization settings, but excluding tactile inputs and tactile-related architecture. During finetuning, we add the same tactile-related architecture as FTP-1 during finetuning and refer to the resulting model as NTP-1. We evaluate NTP-1 on UniVTAC~\cite{chen2026univtac} (seen sensor) and FlexivXense (unseen sensor), using the same finetuning setup as previous section.
\begin{wrapfigure}{r}{0.48\linewidth}
    \centering
    \vspace{-3mm}
    \includegraphics[width=0.95\linewidth]{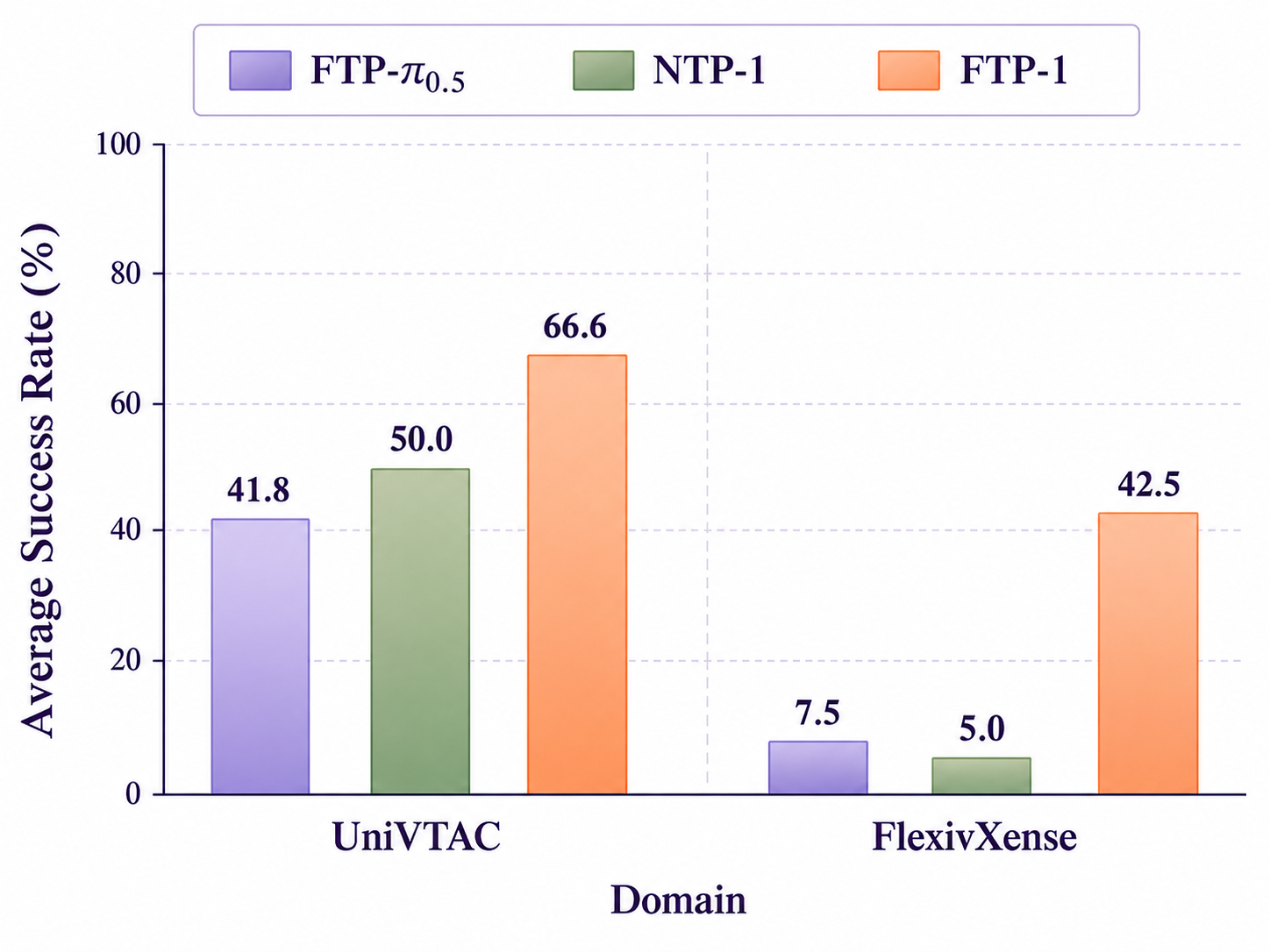}
    \vspace{-3mm}
    \caption{
    Comparison between FTP-1 and NTP-1 on UniVTAC and FlexivXense.
    }
    \label{fig:ntp_experiment_result}
    \vspace{-1em}
\end{wrapfigure}

Results are shown in Fig.~\ref{fig:ntp_experiment_result}, with detailed results provided in App.~\ref{app:ablation_result}. All models use exactly the same architecture, differing only in their pretrained checkpoints. On UniVTAC, NTP-1 outperforms FTP-$\pi_{0.5}$, suggesting that the distribution of the FTP-1-Dataset may be closer to UniVTAC. However, NTP-1 still underperforms FTP-1, indicating that FTP-1 pretraining provides a useful initialization for the tactile branch. On FlexivXense, FTP-1 substantially outperforms NTP-1 by +37.5\%, demonstrating that tactile-based pretraining is essential for transferring to the FlexivXense domain. 
Without tactile-branch pretraining, NTP-1 shows much less robustness to tactile changes and produces unstable actions during key insertion stages. Together, these results support Hypothesis 2 (Transferable Knowledge): the tactile branch of FTP-1 benefits from pretraining, which encodes general tactile manipulation knowledge that can be transferred to downstream contact-rich tasks.

\subsection{Additional Analyses}\label{sec:additional_analyses}

\textbf{Ablation of VLA matched pretraining.} To control for exposure to the robot trajectories used in FTP-1 pretraining, we additionally continue pretraining $\pi_{0.5}$ on the same robot trajectories under the same pretraining budget, using its vision-language-action inputs. After downstream finetuning, this baseline achieves 57.0\% overall success and 44.0\% excluding the two lift tasks on UniVTAC (Tab.~\ref{tab:matched_pretraining}). FTP-1 improves these metrics by 9.7 and 15.5 percentage points, respectively. The matched-data comparison supports the benefit of FTP-1's tactile pretraining beyond the gains from additional robot-trajectory exposure.

\begin{table}[t]
\centering
\caption{Matched-pretraining comparison on UniVTAC. Success rates (\%) after finetuning; both models use the same robot trajectories and pretraining budget.}
\label{tab:matched_pretraining}
\small
\begin{tabular}{lcc}
\toprule
Method & Avg. & Avg. w/o Lift \\
\midrule
$\pi_{0.5}$ with matched pretraining & 57.0 & 44.0 \\
FTP-1 & \textbf{66.7} & \textbf{59.5} \\
\bottomrule
\end{tabular}
\end{table}

\textbf{Ablation of pretraining budget scaling.} We scale the pretraining data and training steps together: the 50\% condition uses half the data and half the steps of the full run. UniVTAC average success increases from 45.2\% without FTP-1 pretraining to 59.0\% at 50\% and 66.7\% at 100\%. These results show consistent gains across the tested budgets; the complete results are reported in App.~\ref{app:budget_scaling}.

\textbf{Tactile feature probing evaluation.} We further evaluate force and slip prediction from frozen tactile features (App.~\ref{app:frozen_probes}). Relative to the T3 initialization, mean attentive slip F1 increases from 0.725 to 0.780 for the FTP-1 trunk and 0.786 for the tokenizer. The force probes also recover quantitative force information. Together, these results provide representation-level evidence for manipulation-relevant contact dynamics that complements the downstream policy evaluations.


\vspace{-1.5mm}
\section{Related Work}
\vspace{-1.5mm}

A detailed review and a comparison of sensor coverage, action generation, and transfer evaluation are provided in App.~\ref{app:related_works} and Tabs.~\ref{tab:tactile_representation_comparison}--\ref{tab:tactile_policy_comparison}. Recent generalist robot policies~\cite{kim2024openvla, black2024pi_0, intelligence2025pi_, bjorck2025gr00t, liu2025rdt} show strong transfer across tasks, embodiments, and environments, but largely omit tactile sensing. Existing tactile policy and Vision-Tactile-Language-Action (VTLA) methods~\cite{heng2025vitacformer, huang2025tactile, cheng2025omnivtla, zhang2026vtla, li2026forcevla2} demonstrate the value of tactile feedback for contact-rich manipulation, but are typically tied to specific sensors, tasks, or embodiments. Tactile representation and policy pretraining methods~\cite{higuera2024sparsh, feng2025anytouch, george2025vital, zhao2024transferable, zheng2026omnivta, wu2025freetacman} improve tactile perception or downstream control, but do not establish an end-to-end sensor-agnostic generalist tactile policy across heterogeneous tactile hardware. 

FTP-1 addresses this gap by extending the generalist policy paradigm to tactile manipulation. Rather than treating tactile feedback as a sensor-specific add-on for policies, FTP-1 studies whether tactile manipulation skills can be pretrained at scale and transferred across heterogeneous sensors and embodiments. This positions FTP-1 as a foundation-policy baseline for tactile manipulation, with evaluation covering both seen and unseen tactile-sensor setups. 

\vspace{-1.5mm}
\section{Conclusion}
\vspace{-1.5mm}

We present FTP-1, a generalist foundation tactile policy pretrained for finetuning transfer across diverse sensors and embodiments. By pretraining on a large-scale heterogeneous tactile manipulation dataset aggregated from 26 sources and covering 21 sensors, FTP-1 learns transferable tactile manipulation skills. Experiments on 5 hardware setups show that FTP-1 improves finetuning performance on both seen and unseen sensors. This provides a shared model-level starting point for tactile policy learning.

\textbf{Limitations and Future Directions.}
As a preliminary exploration of generalist tactile policy learning, FTP-1 mainly focuses on general tactile perception and does not yet address tactile- or force-based servoing and control. Extending our heterogeneous encoding framework to future tactile prediction~\cite{zheng2026omnivta, niu2026learning} and prediction-based low-level control is a promising direction~\cite{xu2026contact, zhang2025kinedex, zhi2025learning}. Another limitation is that the scale and diversity of our pretraining dataset remain limited. Building larger aggregated datasets~\cite{o2024open} and further scaling up pretraining are left for future work.


\clearpage
\acknowledgments{This work is supported by the Shanghai Qi Zhi Institute Innovation Program and the Tsinghua University Dushi Program. We thank Sharpa Pte Ltd. for hardware, computational resources, and the Sharpa North-FTP-1 dataset; Aether Embodied Technology (Beijing) Co., Ltd. for the EgoTac egocentric human dataset and AetherGlove; and PaXini Technology (Shenzhen) Co., Ltd. for the OmniSharingDB human data.}


\bibliography{example,additions/references}  

\clearpage

\appendix

{\Large\bfseries Appendix\par}
\addcontentsline{toc}{section}{Appendix}

\begin{figure}[H]
\centering
  \includegraphics[width=1.0\textwidth]{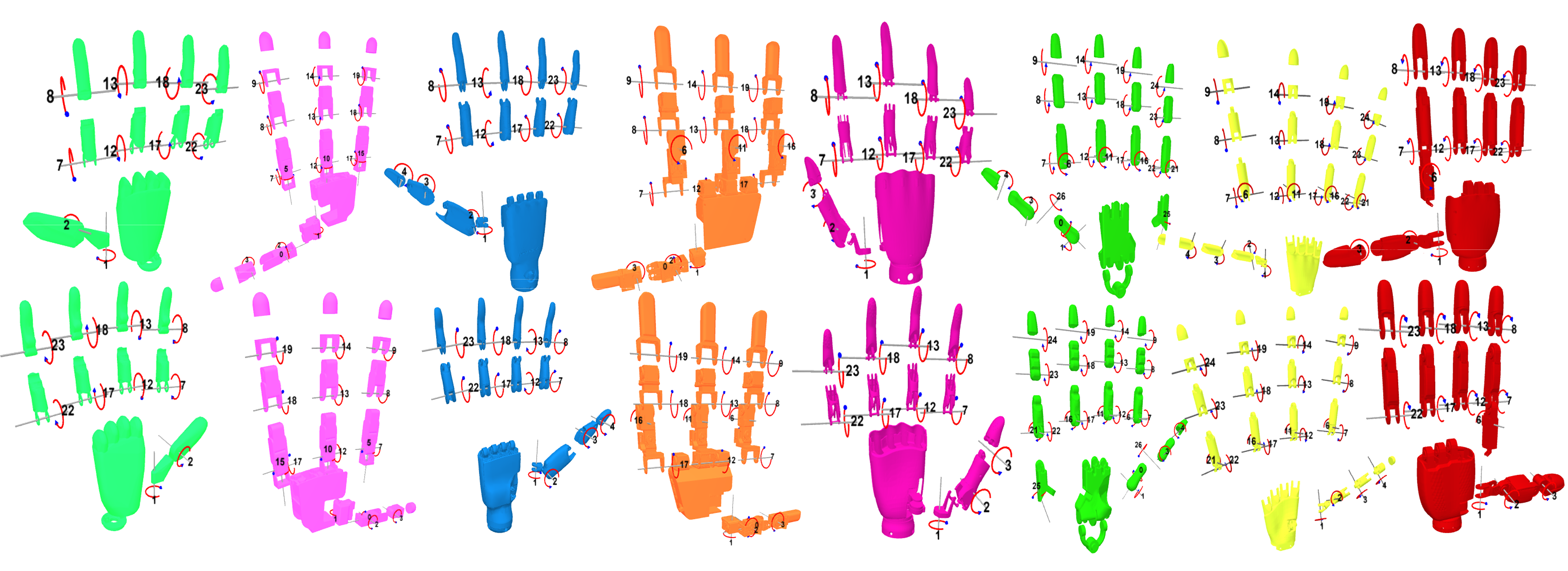}
  \caption{Joint mappings of different robotic hands used in FAAS. From left to right are Ability, Allegro, Inspire, Leap, Oymotion, Shadow, Wuji, and Xhand. The two rows show different views of the joint mappings on the right hand. Refer and copied from the original UniDex~\cite{zhang2026unidex} paper.}
  \label{fig:detailed_faas}
 \vspace*{-10pt}
\end{figure}

\section{Detailed Related Works}\label{app:related_works}

\subsection{Generalist Policy Learning for Robotic Manipulation}

Recent progress in robot learning has shifted from task-specific policies~\cite{chi2025diffusion, zhao2023learning} toward the generalist policy paradigm. Generalist policies~\cite{kim2024openvla, o2024open, barreiros2026careful, intelligence2025pi_, bjorck2025gr00t, pertsch2025fast, liu2025rdt, liu2026rdt2, wang2024scaling, team2025gemini, team2025gemini15} are trained on large-scale robotic datasets~\cite{o2024open, khazatsky2024droid, wu2025robocoin} and can rapidly adapt to new environments, embodiments, and tasks. Beyond robot demonstrations, heterogeneous learning signals such as human motion~\cite{zheng2026egoscale, yang2025egovla, yuan2025motiontrans, cai2025n, yuan2024general, li2025scalable, fu2025metis, zhang2026unidex}, semantic latent tokens~\cite{lyu2026lda, driess2026knowledge, bu2025univla}, video generation~\cite{ye2026world, yuan2026fast, bi2025motus, hu2024video}, and vision-language answering~\cite{intelligence2025pi_, lin2025onetwovla, lin2026systematic} have also been incorporated into training frameworks to improve generalization. However, existing generalist policies largely omit tactile inputs, which are crucial for contact-rich manipulation.

\subsection{Tactile Policy Learning for Contact-Rich Manipulation}

The absence of generalist tactile policies is largely due to the heterogeneity of tactile hardware~\cite{cheng2026taco}. Across platforms, tactile observations vary in modality, making cross-sensor generalization challenging~\cite{higuera2024sparsh, wi2026tactalign, jia2026feel}. Nevertheless, recent sensor-specific visuo-tactile policy learning works~\cite{heng2025vitacformer, huang20243d, niu2026learning, xue2025reactive, huang2025spatially, yu2023mimictouch, fang2026force, he2025foar, tian2026vitas, choi2026wild, liu2025vitamin, li2025vitaminb, xu2025exumi, higuera2025tactile} show that tactile signals substantially improve contact-rich manipulation. In parallel, Vision-Tactile-Language-Action (VTLA) models~\cite{huang2025tactile, bi2025vla, zhang2026vtla, zhang2026tacvla, yu2026forcevla, li2026forcevla2} augment pretrained vision-based foundation policies with tactile inputs~\cite{black2024pi_0, wu2026pragmatic}. In contrast, FTP-1 is a natively tactile foundation model~\cite{generalist2025gen0}, rather than a task-specific policy or a vision-policy extension.

\subsection{Large-Scale Tactile Representation Pretraining}

Recent works have explored tactile representation pretraining to improve tactile perception for downstream tasks. One line of work~\cite{liu2025vtdexmanip, liu2024masked, cheng2025touch100k, zhu2026touch, chen2026univtac, zhao2024transferable}, including Sparsh~\cite{higuera2024sparsh}, AnyTouch~\cite{feng2025anytouch}, and ViTaL~\cite{george2025vital}, learns tactile encoders from vision-tactile pairs for downstream policy finetuning. However, these representations are not optimized end-to-end for robotic manipulation. Another line directly pretrains tactile policies on action-labeled tactile manipulation data, including VTAO-BiManip~\cite{sun2025vtao}, FreeTacMan~\cite{wu2025freetacman}, OmniVTA~\cite{zheng2026omnivta}, and TAMEn~\cite{wu2026tamen}, but remains limited by sensor-specific setups and dataset diversity and scale. FTP-1 instead offers a sensor-agnostic generalist foundation policy with large-scale pretraining for diverse tactile robot setups.

\subsection{Comparison with Tactile Representation and Policy Models}
\label{app:tactile_model_comparison}

Tables~\ref{tab:tactile_representation_comparison} and~\ref{tab:tactile_policy_comparison} compare representative tactile models along complementary axes. Sensor coverage refers to the reported tactile dataset or training setup, rather than the number of physical sensor units mounted on a robot. We distinguish transfer between sensor instances, transfer between sensor designs, and downstream action generation. In particular, evaluating a pretrained encoder with a task-specific policy head is different from pretraining a shared action-generating tactile policy. ``NR'' means that the corresponding evaluation or count is not reported in the cited work; it does not imply that the method cannot support that setting.

\begin{table}[t]
\centering
\caption{Tactile coverage and sensor-transfer evaluation in representative models. Counts follow each paper's stated dataset or experimental setting. Sensor-transfer protocols are listed explicitly because perception probes, calibration, and policy finetuning measure different capabilities.}
\label{tab:tactile_representation_comparison}
\footnotesize
\setlength{\tabcolsep}{3pt}
\renewcommand{\arraystretch}{1.14}
\begin{tabular}{@{}p{0.14\linewidth}p{0.27\linewidth}p{0.20\linewidth}p{0.33\linewidth}@{}}
\toprule
\textbf{Model} & \textbf{Reported tactile coverage} & \textbf{Learning target} & \textbf{Unseen-sensor evaluation} \\
\midrule
T3~\cite{zhao2024transferable}
& 13 sensors in FoTa; image inputs
& Shared tactile representation
& GelSight Svelte and DenseTact~2.0 held out in transfer experiments; classification/pose estimation, with and without finetuning \\
\addlinespace[2pt]
Sparsh~\cite{higuera2024sparsh}
& 3 families: DIGIT, GelSight~2017, GelSight Mini
& Self-supervised tactile representation
& New physical instances of covered families; frozen-feature perception tasks and downstream policy learning \\
\addlinespace[2pt]
AnyTouch~\cite{feng2025anytouch}
& 5 training sensor types; image/video inputs
& Static/dynamic tactile representation
& TACTO and Taxim synthetic sensors; material classification using linear probes \\
\addlinespace[2pt]
FEATS~\cite{helmut2024feats}
& 1 type: GelSight Mini; multiple units/gels
& Dense 3D force distribution
& New GelSight Mini unit/gel combinations; force estimation, with additional multi-unit training also evaluated \\
\addlinespace[2pt]
FeelAnyForce \cite{shahidzadeh2025feelanyforce}
& GelSight Mini training; Mini and DIGIT evaluation
& 3D contact force and depth
& Additional Mini units/gels, with and without calibration; DIGIT with regressor-head finetuning \\
\addlinespace[2pt]
Tactile-VLA \cite{huang2025tactile}
& Dual force-sensing fingertips; sensor-type count NR
& Position/force actions and tactile reasoning
& NR; reported generalization concerns instructions, objects, and task conditions \\
\addlinespace[2pt]
OmniVTLA \cite{cheng2025omnivtla}
& Paxini Gen2 in ObjTac and robot setups; additional encoder-training datasets (total type count NR)
& Aligned tactile features and actions
& Held-out sensor policy evaluation NR \\
\addlinespace[2pt]
OmniVTA~\cite{zheng2026omnivta}
& 4 sensor types in dataset; 3 in tactile-VAE validation; Xense in robot policy evaluation
& Tactile dynamics and actions
& Held-out sensor policy evaluation NR \\
\addlinespace[2pt]
\textbf{FTP-1}
& 21 pretraining sensors; image, array, and state inputs
& Shared tactile action-generating policy
& Policy finetuning on Xense and Contactile, absent from FTP-1 pretraining \\
\bottomrule
\end{tabular}
\end{table}

\begin{table}[t]
\centering
\caption{Action generation, embodiment coverage, and real-robot evaluation. ``Policy setups'' counts the downstream robot configurations evaluated in the cited work, not all devices used to collect pretraining data. A downstream policy head uses a pretrained representation but is trained for the manipulation task. Cross-embodiment entries describe the reported protocol rather than equating multiple hardware evaluations with a held-out-embodiment transfer test.}
\label{tab:tactile_policy_comparison}
\footnotesize
\setlength{\tabcolsep}{3pt}
\renewcommand{\arraystretch}{1.14}
\begin{tabular}{@{}p{0.14\linewidth}p{0.21\linewidth}p{0.20\linewidth}p{0.14\linewidth}p{0.24\linewidth}@{}}
\toprule
\textbf{Model} & \textbf{Action generation} & \textbf{Policy setups} & \textbf{Real-robot policy evaluation} & \textbf{Cross-embodiment evaluation} \\
\midrule
T3~\cite{zhao2024transferable}
& Downstream behavior-cloning head
& 1: Franka Panda + parallel gripper
& Yes; insertion
& NR \\
\addlinespace[2pt]
Sparsh~\cite{higuera2024sparsh}
& Downstream diffusion policy
& 1: Franka + tactile hand
& Yes; bead maze
& NR \\
\addlinespace[2pt]
AnyTouch~\cite{feng2025anytouch}
& Downstream imitation policy
& 1: xArm~6 + Robotiq gripper
& Yes; pouring
& NR \\
\addlinespace[2pt]
FEATS~\cite{helmut2024feats}
& Force estimator
& NR
& NR
& NR \\
\addlinespace[2pt]
FeelAnyForce \cite{shahidzadeh2025feelanyforce}
& Force estimator used by a feedback controller
& No learned policy; physical force-feedback tasks
& Feedback-control demonstrations
& NR \\
\addlinespace[2pt]
Tactile-VLA \cite{huang2025tactile}
& End-to-end position/force policy
& 1 reported robot gripper setup; UMI data collection
& Yes
& UMI-to-robot data use; held-out robot test NR \\
\addlinespace[2pt]
OmniVTLA \cite{cheng2025omnivtla}
& End-to-end action policy
& 2: UR5 + gripper / dexterous hand
& Yes
& Shared encoder evaluated with separate gripper/hand policy finetuning \\
\addlinespace[2pt]
OmniVTA~\cite{zheng2026omnivta}
& Diffusion policy + tactile controller
& 1: xArm~7 + parallel gripper
& Yes
& Robot and TacUMI collection with identical grippers; held-out robot test NR \\
\addlinespace[2pt]
\textbf{FTP-1}
& Shared end-to-end action policy
& 5: 1 simulation and 4 real setups
& Yes
& Shared policy finetuning across grippers and dexterous hands, including new hardware setups \\
\bottomrule
\end{tabular}
\end{table}

\paragraph{Tactile representations and physically grounded estimation.}
T3~\cite{zhao2024transferable} and AnyTouch~\cite{feng2025anytouch} explicitly evaluate transfer across sensor designs, while Sparsh~\cite{higuera2024sparsh} evaluates representations on new physical sensor instances and diverse downstream tasks. These works also demonstrate the use of tactile representations in real-robot policy learning. FEATS~\cite{helmut2024feats} and FeelAnyForce \cite{shahidzadeh2025feelanyforce} provide complementary physically grounded supervision: FEATS learns spatial force distributions from finite-element labels, whereas FeelAnyForce jointly predicts contact force and depth and studies adaptation across sensor units and designs. Their sensor-transfer results establish relevant precedents for tactile perception; FTP-1 studies action-generating policy pretraining and subsequent manipulation-policy finetuning across heterogeneous tactile hardware.

\paragraph{Tactile action-generating models.}
Tactile-VLA \cite{huang2025tactile} connects tactile feedback to position--force action generation and language-conditioned adaptation. OmniVTLA \cite{cheng2025omnivtla} aligns tactile features with vision and language and evaluates gripper and dexterous-hand policies. OmniVTA~\cite{zheng2026omnivta} combines visuo-tactile world modeling, a diffusion policy, and a reactive tactile controller; its dataset supports four sensor types, while its reported real-robot policy experiments use Xense. FTP-1 combines heterogeneous tactile inputs with a shared pretrained action-generating policy and evaluates finetuning across five hardware setups, including two tactile sensors absent from pretraining. This comparison separates the breadth of a data collection or representation framework from the sensor and embodiment transfer actually evaluated at the policy level.

\section{Details of FTP-1 Architecture}\label{app:architecture}

\subsection{Unified Action Space (UAS)}\label{app:UAS}

Different embodiments use different action spaces for robotic control. For example, one robot may control a single arm in joint space, while another may require bimanual end-effector pose control. To handle this heterogeneity, FTP-1 adopts a Unified Action Space (UAS) that represents all robot commands as fixed-length sparse vectors~\cite{liu2025rdt}. For one control step, the predicted action is
$\mathbf{a}=[\mathbf{a}^{L}, \mathbf{a}^{R}, \mathbf{a}^{ego}, \mathbf{a}^{sup}] \in \mathbb{R}^{D}$
where $\mathbf{a}^{L}$ and $\mathbf{a}^{R}$ denote the left- and right-arm control signals, $\mathbf{a}^{ego}$ denotes the head-pose control signal, and $\mathbf{a}^{sup}$ provides supplementary slots for additional controls such as locomotion or waist motion.

For each arm $b\in\{L,R\}$, the arm action is defined as
$\mathbf{a}^{b}=[\mathbf{t}_{w}^{b}, \mathbf{r}_{w}^{b}, \mathbf{q}_{\mathrm{arm}}^{b}, \mathbf{q}_{\mathrm{hand}}^{b}]$,
where $\mathbf{t}_{w}^{b}\in\mathbb{R}^{3}$ is the wrist translation, $\mathbf{r}_{w}^{b}\in\mathbb{R}^{6}$ is the 6D wrist rotation representation~\cite{chi2024universal}, $\mathbf{q}_{\mathrm{arm}}^{b}\in\mathbb{R}^{7}$ denotes arm joints, and $\mathbf{q}_{\mathrm{hand}}^{b}\in\mathbb{R}^{32}$ denotes canonical hand joint control slots.
\begin{wrapfigure}{r}{0.5\linewidth}
    \centering
    \vspace{-0.6em}
    \includegraphics[width=1.0\linewidth]{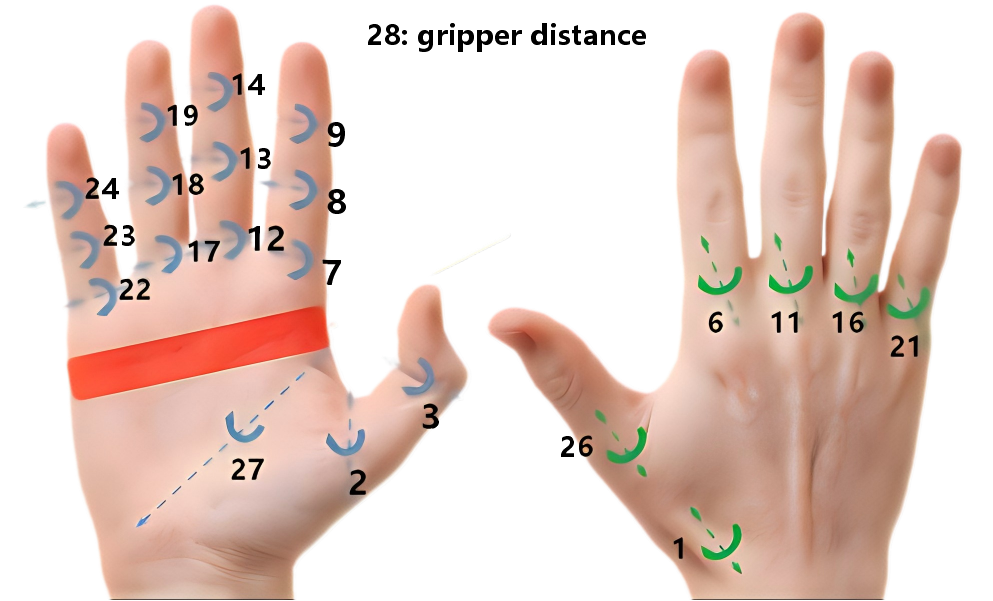}
    \caption{\small FAAS definition for the human hand.}
    \label{fig:faas_human_hand}
    \vspace{-1.0em}
\end{wrapfigure}
For $\mathbf{q}_{\mathrm{hand}}^{b}$, following UniDex~\cite{zhang2026unidex}, we use a Function--Actuator--Aligned Space (FAAS) to unify the control action spaces of different end-effectors.

\textbf{Brief Review of FAAS~\cite{zhang2026unidex}}. For the hand joint action space $q_{hand}$, we follow UniDex~\cite{zhang2026unidex} and adopt the Function–Actuator–Aligned Space (FAAS), which unifies the action spaces of different end-effectors by assigning functionally similar joints to the same action slot. Fig.~\ref{fig:detailed_faas} shows the FAAS definitions for commonly used dexterous hands, reproduced from the original paper. By providing a universal action interface, FAAS enables new end-effectors to be integrated by simply defining their mappings to the predefined functional slots, thereby sharing the same action semantics and knowledge as those used during pretraining.

We further define the FAAS for the human hand in Fig.~\ref{fig:faas_human_hand}. For parallel grippers, we use an independent slot 28. For head control, $\mathbf{a}^{ego}=[\mathbf{t}^{ego}, \mathbf{r}^{ego}]$, where $\mathbf{t}^{ego}\in\mathbb{R}^{3}$ and $\mathbf{r}^{ego}\in\mathbb{R}^{6}$ denote the translation and rotation of the head pose.

For each embodiment, only the supported control slots are filled in the UAS. During training, we apply a mask $\mathbf{M}\in\{0,1\}^{D}$ to exclude missing or unsupported action dimensions from the loss. With the UAS, FTP-1 always predicts actions in the same layout, enabling unified training across embodiments with different action spaces.

\begingroup
\raggedbottom
\paragraph{FAAS mapping for the Sharpa Wave hand.}
The Sharpa Wave hand used in the Sharpa North and Sharpa\&Dexmate setups has 22 actuated joints per hand. We map these joints into the 32-dimensional canonical hand vector $\mathbf{q}_{\mathrm{hand}}^{b}$ using the correspondence in Tab.~\ref{tab:sharpa_hand_faas}; Fig.~\ref{fig:sharpa_hand_faas} visualizes this mapping on the hand. All indices are zero-based and local to one hand; the same mapping applies independently to the left and right hands. In the native joint order listed in the table, the FAAS slot sequence is
\[
\mathbf{s} = [1,2,3,26,4,\;7,6,8,9,\;12,11,13,14,\;17,16,18,19,\;25,22,21,23,24].
\]
For a native hand-joint vector $\mathbf{h}^{b}\in\mathbb{R}^{22}$, we assign $q_{\mathrm{hand},s_k}^{b}=h_k^{b}$ for $k=0,\ldots,21$. In particular, slot 25 represents the little-finger carpometacarpal (CMC) joint, and slot 26 represents thumb metacarpophalangeal (MCP) abduction/adduction. The remaining slots $\{0,5,10,15,20,27,28,29,30,31\}$ are unused for this hand, are zero-filled when constructing the canonical vector, and are excluded from the action loss. The same correspondence is used for proprioceptive joint inputs and action targets. At deployment, the 22 joint commands are recovered by selecting the occupied slots in the native joint order, with the configured joint-representation conversion applied.

\begin{figure}[H]
    \centering
    \includegraphics[trim=0 0 0 25bp,clip,width=\linewidth]{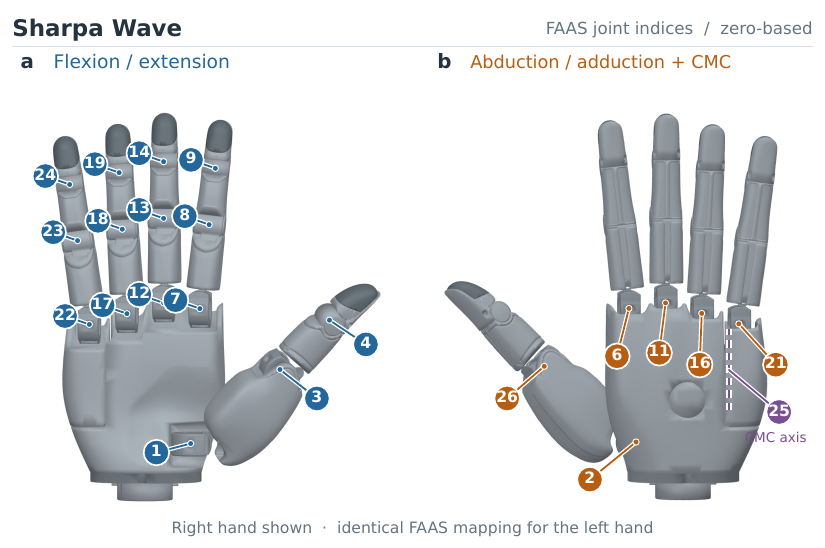}
    \caption{FAAS indices on the Sharpa Wave right hand. (a) Flexion/extension (blue). (b) Abduction/adduction (orange) and little-finger CMC (purple dashed axis). Dots mark the other joint origins; numbers denote zero-based FAAS slots. Both views use the robot model. The left hand uses the same mapping.}
    \label{fig:sharpa_hand_faas}
\end{figure}

\begin{table}[H]
    \centering
    \caption{Mapping from native Sharpa Wave hand joints to FAAS slots. Both index columns are zero-based. CMC: carpometacarpal; MCP: metacarpophalangeal; IP: interphalangeal; PIP/DIP: proximal/distal interphalangeal; FE: flexion/extension; AA: abduction/adduction. Little finger corresponds to \texttt{pinky} in the robot joint names.}
    \label{tab:sharpa_hand_faas}
    \small
    \renewcommand{\arraystretch}{1.0}
    \begin{tabular}{rllr}
        \toprule
        Native index & Finger & Joint / degree of freedom & FAAS slot \\
        \midrule
        0  & Thumb  & CMC FE & 1  \\
        1  & Thumb  & CMC AA & 2  \\
        2  & Thumb  & MCP FE & 3  \\
        3  & Thumb  & MCP AA & 26 \\
        4  & Thumb  & IP     & 4  \\
        \midrule
        5  & Index  & MCP FE & 7  \\
        6  & Index  & MCP AA & 6  \\
        7  & Index  & PIP    & 8  \\
        8  & Index  & DIP    & 9  \\
        \midrule
        9  & Middle & MCP FE & 12 \\
        10 & Middle & MCP AA & 11 \\
        11 & Middle & PIP    & 13 \\
        12 & Middle & DIP    & 14 \\
        \midrule
        13 & Ring   & MCP FE & 17 \\
        14 & Ring   & MCP AA & 16 \\
        15 & Ring   & PIP    & 18 \\
        16 & Ring   & DIP    & 19 \\
        \midrule
        17 & Little & CMC    & 25 \\
        18 & Little & MCP FE & 22 \\
        19 & Little & MCP AA & 21 \\
        20 & Little & PIP    & 23 \\
        21 & Little & DIP    & 24 \\
        \bottomrule
    \end{tabular}
\end{table}
\par
\endgroup

\subsection{Details of Heterogeneous Tactile Encoders}\label{app:tactile_encoder}

Here we provide the detailed architecture of tactile encoders for each type of tactile input.

\textbf{Image-type inputs.} For inputs of image shape \((H,W,C)\), we follow T3-Encoder~\cite{zhao2024transferable}: each tactile image is resized to \(224 \times 224\), encoded by a sensor-specific Transformer encoder (depth=3, width=768, head=12), and then passed to a shared Transformer module (depth=9, width=768, head=12) for unified tokenization. The shared module is initialized from pretrained T3 weights~\cite{zhao2024transferable}, and the final \texttt{[CLS]} token is used as the tactile representation.

\textbf{Array-type inputs.} For inputs of shape \((H,W,D)\), where \(H,W\) is resolution of the array and $D$ is the dimension of signal for each unit, we apply Fourier encoding~\cite{huang2025spatially} on the signal dimension, concatenate it with the original input, and process the result with a 3-layer CNN~\cite{wu2017introduction} followed by a 2-layer ReLU MLP to obtain the tactile token.

\textbf{State-type inputs.} For inputs of shape \((D,)\), we apply Fourier encoding~\cite{huang2025spatially}, concatenate it with the original state, and use a 3-layer ReLU MLP to obtain the tactile token.

All tactile tokens are then normalized by LayerNorm~\cite{ba2016layer}, augmented with shared function-area embeddings, and projected by a 2-layer GELU MLP to match the tactile expert input dimension.

\subsection{Adaptive RMSNorm Proprioception Injector}\label{app:proprioception_injector}

In preliminary experiments, we found that injecting proprioception through adaptive RMSNorm~\cite{intelligence2025pi_} improves generalization and robustness compared to using an independent proprioceptive token in the Transformer expert. We Fourier-encode~\cite{huang2025spatially} the proprioceptive state and process it with a 3-layer MLP. The resulting features are LayerNorm-normalized~\cite{ba2016layer}, concatenated with layer-normalized flow-matching timestep features from \(\pi_{0.5}\)~\cite{intelligence2025pi_}, and injected into the attention blocks via adaptive RMSNorm~\cite{intelligence2025pi_, peebles2023scalable}.

\subsection{Attention Connectivity}\label{app:attention_connectivity}

The three experts have separate attention projections and feed-forward parameters. At each layer, their query, key, and value sequences are concatenated for a shared masked attention operation, and each expert then applies its own output projection and feed-forward block. With query streams as rows and key streams as columns, both ordered as vision-language, tactile, and action, the block visibility mask is
\[
\mathbf{M}_{\mathrm{expert}} =
\left[\begin{array}{ccc}
1 & 0 & 0 \\
0 & 1 & 0 \\
1 & 1 & 1
\end{array}\right].
\]
An entry of one permits attention, with input-padding masks applied in addition. The action expert can therefore use both perceptual streams when predicting an action chunk, while the two perceptual experts process their respective observations independently.

\subsection{Other Details}\label{app:other_architecture}

For state and action normalization, we adopt z-score normalization~\cite{black2024pi_0}, which yielded better action generation quality than quantile-based normalization~\cite{intelligence2025pi_} in our experiments, especially for fine-grained action in contact-rich manipulation. The tactile expert is a Transformer with width 1024, depth 18, MLP dimension 4096, 8 attention heads, and head dimension 256. We train the model with AdamW~\cite{loshchilov2017decoupled}. We also try Muon~\cite{liu2025muon} optimizer in our earlier experiments. Although we find that Muon optimizer improves convergence speed and offline action MSE significantly, it reduces generalization and robustness in real-robot rollouts, so we use AdamW in the final model.

\section{Details of FTP-1 Pretraining Datasets}\label{app:pretrain_dataset_and_sampling}

\begin{table}[t]
    \centering
    \scriptsize
    \setlength{\tabcolsep}{2.8pt}
    \renewcommand{\arraystretch}{1.08}
    \caption{Overview of the FTP-1 pretraining datasets and sampling ratios. Data sources are grouped into three domains according to their end-effector types: human hand, dexterous hand, and gripper. To address data imbalance across sources, we resample data from each source according to the sampling scale shown in the table.}
    \label{tab:pretrain_data_mixture}
    \begin{tabular*}{\columnwidth}{@{\extracolsep{\fill}}p{2.20cm}p{1.55cm}p{4.20cm}rr@{}}
        \toprule
        Data Source & Embodiment & Tactile / Force Sensor(s) (type) & Sample-Scale & Final-Percentage \\
        \midrule
        \multicolumn{5}{l}{\textbf{Human Hand Domains (20\%)}} \\
        \midrule
        EgoTac-Open & Human & AetherGlove (array) & 0.89 & 8.2\% \\
        EgoTac-Kitchen & Human & AetherGlove (array) & 1.63 & 5.5\% \\
        EgoTac-Hotel & Human & AetherGlove (array) & 2.82 & 3.8\% \\
        OpenTouch~\cite{song2025opentouch} & Human & OpenTouchGlove (array) & 11.09 & 1.5\% \\
        OmniSharingDB & Human & PaxiniGlove (state) & 21.65 & 1.0\% \\
        \midrule
        \multicolumn{5}{l}{\textbf{Dexterous Hand Domains (30\%)}} \\
        \midrule
        Sharpa North-FTP-1 & DexHand & Sharpa DTC (image) & 3.29 & 9.0\% \\
        HumanoidEveryday\allowbreak{} \cite{zhao2025humanoid} & DexHand & UnitreeDex3 (state) & 13.97 & 6.0\% \\
        OpenLETDex & DexHand & LinkerHandL6Touch (array) & 14.14 & 6.0\% \\
        MotionTrans~\cite{yuan2025motiontrans} & DexHand & InspireHand (state) & 26.39 & 3.6\% \\
        DexumiInspire~\cite{xu2025dexumi} & UMI+DexHand & DexumiInspire (state) & 35.62 & 3.0\% \\
        DexumiXHand~\cite{xu2025dexumi} & UMI+DexHand & DexumiXHand (state) & 56.02 & 2.4\% \\
        \midrule
        \multicolumn{5}{l}{\textbf{Gripper / UMI Domains (50\%)}} \\
        \midrule
        VTouch-D-WHEEL & Gripper & ViTai-GF (image) & 2.77 & 5.7\% \\
        FreeTacMan~\cite{wu2025freetacman} & UMI & FreeTacMan (image) & 7.75 & 4.4\% \\
        TouchInTheWild~\cite{zhu2026touch} & UMI & 3DViTac (array) & 10.79 & 4.0\% \\
        VTouch-QINGLONG & Gripper & ViTai-GF (image) & 12.73 & 3.9\% \\
        exUMI~\cite{xu2025exumi} & UMI & exUMI (image) & 13.73 & 3.8\% \\
        RH20T-Cfg6~\cite{fang2023rh20t} & Gripper & ATIAxia80M20 (state) & 13.88 & 3.8\% \\
        ViTaMIn~\cite{liu2025vitamin} & UMI & ViTaMIn (image) & 22.19 & 3.4\% \\
        RH20T-Cfg7~\cite{fang2023rh20t} & Gripper & uSkin (array); ATIAxia80M20 (state) & 22.89 & 3.4\% \\
        RH20T-Cfg5~\cite{fang2023rh20t} & Gripper & FrankaTorque (state) & 24.42 & 3.3\% \\
        REASSEMBLE~\cite{sliwowski2025reassemble} & Gripper & AIDIN-FT (state) & 27.77 & 3.2\% \\
        Unit-Bimanual~\cite{xu2025unit} & Gripper & GelSight-Mini (image) & 63.93 & 2.6\% \\
        RDP~\cite{xue2025reactive} & Gripper & GelSight-Mini (image); MCTac (image); FlexivGripperTorque (state) & 89.76 & 2.4\% \\
        Unit~\cite{xu2025unit} & Gripper & GelSight-Mini (image) & 132.83 & 2.2\% \\
        VLA-Touch~\cite{bi2025vla} & Gripper & GelSight-Mini (image) & 170.36 & 2.0\% \\
        RDP-Bimanual~\cite{xue2025reactive} & Gripper & GelSight-Mini (image); MCTac (image); FlexivGripperTorque (state) & 199.20 & 1.9\% \\
        \bottomrule
    \end{tabular*}
\end{table}

In this section, we describe the FTP-1 pretraining datasets. An overview is provided in Tab.~\ref{tab:pretrain_data_mixture}. The dataset is aggregated from 26 data sources, covering human demonstrations, dexterous-hand data, parallel-gripper data, and UMI-style data~\cite{chi2024universal}. 
These sources include large-scale robotic manipulation datasets~\cite{fang2023rh20t, zhao2025humanoid, wu2025freetacman, yuan2025motiontrans,zhu2026touch}, human manipulation datasets with tactile annotations~\cite{song2025opentouch} and smaller-scale experimental datasets~\cite{xu2025dexumi,xu2025exumi,liu2025vitamin,sliwowski2025reassemble,xu2025unit,xue2025reactive,bi2025vla}. In addition, we collect a new robotic dataset, Sharpa North-FTP-1, using the Sharpa North hardware platform~\cite{heng2025vitacformer}, containing approximately 4,000 long-horizon dexterous manipulation demonstrations.

FTP-1-Dataset spans 21 tactile sensors, including 7 image-type sensors (Sharpa DTC, ViTai-GF, FreeTacMan, exUMI, ViTaMIn, GelSight-Mini, MCTac), 5 array-type sensors (AetherGlove, OpenTouchGlove, LinkerHandL6Touch, 3DViTac, uSkin), and 9 state-type sensors (PaxiniGlove, UnitreeDex3, InspireHand, DexumiInspire, DexumiXHand, ATI Axia80-M20, FrankaTorque, AIDIN-FT, FlexivGripperTorque). For data sources with wrist-pose or head-pose annotations, we transform the poses into a unified coordinate direction definition. For language annotations, we use GPT-4o~\cite{hurst2024gpt} to rewrite the original task instructions to increase linguistic diversity. 

Since data sources vary substantially in scale, we adjust the sampling ratio of each source according to the scale in Tab.~\ref{tab:pretrain_data_mixture}. After resampling, the mixture contains approximately 20\% human-hand data, 30\% dexterous-hand data, and 50\% gripper data. In total, the dataset comprises around 3,000 hours of tactile-based manipulation data, covering diverse contact conditions, tasks, scenes, and embodiments. During pretraining, normalization statistics are computed independently for each dataset. Our heterogeneous tactile encoders are organized by tactile sensor rather than data source; therefore, data from the same sensor share the same tactile encoder across different data sources. Pretraining on this large-scale heterogeneous dataset enables FTP-1 to learn transferable tactile manipulation knowledge that adapts to diverse tactile sensors and embodiments.

\section{Training Details}\label{app:training_details}

\textbf{Pretraining Setting.} 
FTP-1 is built on the $\pi_{0.5}$ codebase~\cite{intelligence2025pi_}. We initialize the vision encoder, language tokenizer, vision-language expert, and action expert from $\pi_{0.5}$. The shared image-tactile Transformer trunk is initialized from T3~\cite{zhao2024transferable}. Sensor-specific tactile encoders, the tactile expert, functional-area embeddings, the adaptive RMSNorm proprioception injector, and the action projector are initialized from scratch. Pretraining is performed on 48 NVIDIA H20 GPUs for 50k steps with global batch size 768 and a learning rate decayed from $1\times10^{-4}$ to $5\times10^{-5}$. Performance saturates beyond 50k steps, likely due to limited tactile data diversity and the trade-off between preserving $\pi_{0.5}$ knowledge and learning tactile skills. Further increasing dataset diversity or co-training with $\pi_{0.5}$ datasets may lead to improved scaling behavior and we leave it to future works.

\textbf{Pretraining Infrastructure.} 
To support large-scale heterogeneous pretraining, we develop a new training infrastructure that automatically assigns data from different domains to separate GPUs. This ensures that samples within each GPU batch share the same data format, enabling efficient parallel training for large-scale heterogeneous datasets. Gradients of domain-specific modules are updated independently, while gradients of shared modules are merged before the joint update. 

\textbf{Finetuning Setting.} 
Downstream finetuning uses 8 NVIDIA A800 GPUs for 20k steps per dataset, with batch size 64 and learning rate decayed from $5\times10^{-5}$ to $5\times10^{-6}$. Other settings are the same as in pretraining.

\begin{table}[t]
    \centering
    \caption{Overview of the evaluation setups. Components marked as \textbf{unseen} are not covered by FTP-1 large-scale heterogeneous pretraining.}
    \vspace{-1mm}
    \label{tab:evaluation_hardware_setups}
    \scriptsize
    \setlength{\tabcolsep}{3pt}
    \begin{tabular}{lccccc}
        \toprule
        \textbf{Setup Name} & \textbf{Domain} & \textbf{Robot} & \textbf{End-effector} & \textbf{Tactile Sensor} & \textbf{Sensor Type} \\
        \midrule
        UniVTAC       & Sim~\cite{chen2026univtac} \textbf{(unseen)}, Gripper  & Franka & Franka & GelSight-Mini~\cite{yuan2017gelsight} & image \\
        Sharpa North    & Real, Dexterous Hand & Sharpa North & Sharpa Wave & Sharpa DTC~\cite{heng2025vitacformer} & image \\
        Sharpa\&Dexmate & Real, Dexterous Hand & Dexmate \textbf{(unseen)} & Sharpa Wave & Sharpa DTC~\cite{heng2025vitacformer} & image\\
        \midrule
        FlexivXense   & Real, Gripper & Flexiv & Robotiq & Xense \textbf{(unseen)} & image\\
        TactileUMI    & Real, Gripper & Franka & 3D-printed~\cite{huang2025tactile} \textbf{(unseen)} & Contactile~\cite{velasco2025touch} \textbf{(unseen)} & array \\
        \bottomrule
    \end{tabular}
\end{table}

\section{Details of Evaluation}\label{app:evaluation setup}

\subsection{Hardware Setups}\label{app:hardware_setup}

The detailed hardware setups are provided in Tab~\ref{tab:evaluation_hardware_setups}.

\subsection{Evaluation Setups of Seen Sensors}\label{app:evaluation_setup_seen_sensor}

We evaluate FTP-1 on three embodiments: UniVTAC~\cite{chen2026univtac} (simulation), and Sharpa North and Sharpa\&Dexmate. These setups cover 2 tactile sensors, GelSight-Mini~\cite{yuan2017gelsight} and Sharpa DTC~\cite{heng2025vitacformer}; hardware details are summarized in App.~\ref{app:hardware_setup}. Since these sensors are included in pretraining, both the tactile tokenizer and tactile expert are initialized from FTP-1 pretrained checkpoints.

For the UniVTAC simulation benchmark~\cite{chen2026univtac}, we follow the original protocol and finetune each task with 50 official released demonstrations. We exclude two tasks that are unsuitable for evaluation. We exclude ``Grasp Classify", as it has already been solved perfectly by previous baselines. We also exclude ``Insert HDMI" task because the motion-planning-based data collection in the original benchmark leads to perfect insertion in every demonstration, providing insufficient contact-feedback for evaluating tactile policy learning. The remaining 6 tasks cover two types of contact-rich behavior: in-hand manipulation (Lift Bottle, Lift Can, Put Bottle) and contact-aware insertion/extraction (Pull-out Key, Insert Hole, Insert Tube).

For Sharpa North, we finetune FTP-1 on 250-400 demonstrations across three long-horizon tasks: drawing a smiley face on a balloon (Draw Balloon), attaching a finger to a small damaged model hand (Fix Hand), and twisting a bottle cap with two hands (Twist Cap). For Fix Hand, we evaluate the task in two stages: first tearing off the small sticker from the damaged hand (Tear), and then attaching the finger to complete the repair (Finish). These tasks are deliberately challenging and are designed to probe the capability boundary of FTP-1, requiring deformable-object interaction, fine-grained contact control, small-object manipulation, bimanual coordination, and other complex contact-rich skills.

For Sharpa\&Dexmate, we include two additional tasks: flipping a book (Flip Book) and wiping a dish (Wipe Dish). These tasks complement the evaluation suite by testing contact-rich pressing and force-control behaviors. Each task is finetuned with 100 demonstrations.

\subsection{Visualization of Policy Rollouts for Seen-Sensor Evaluation}
\label{app:seen_sensor_rollout}

\begin{figure*}[t]
  \centering
  \includegraphics[width=\linewidth]{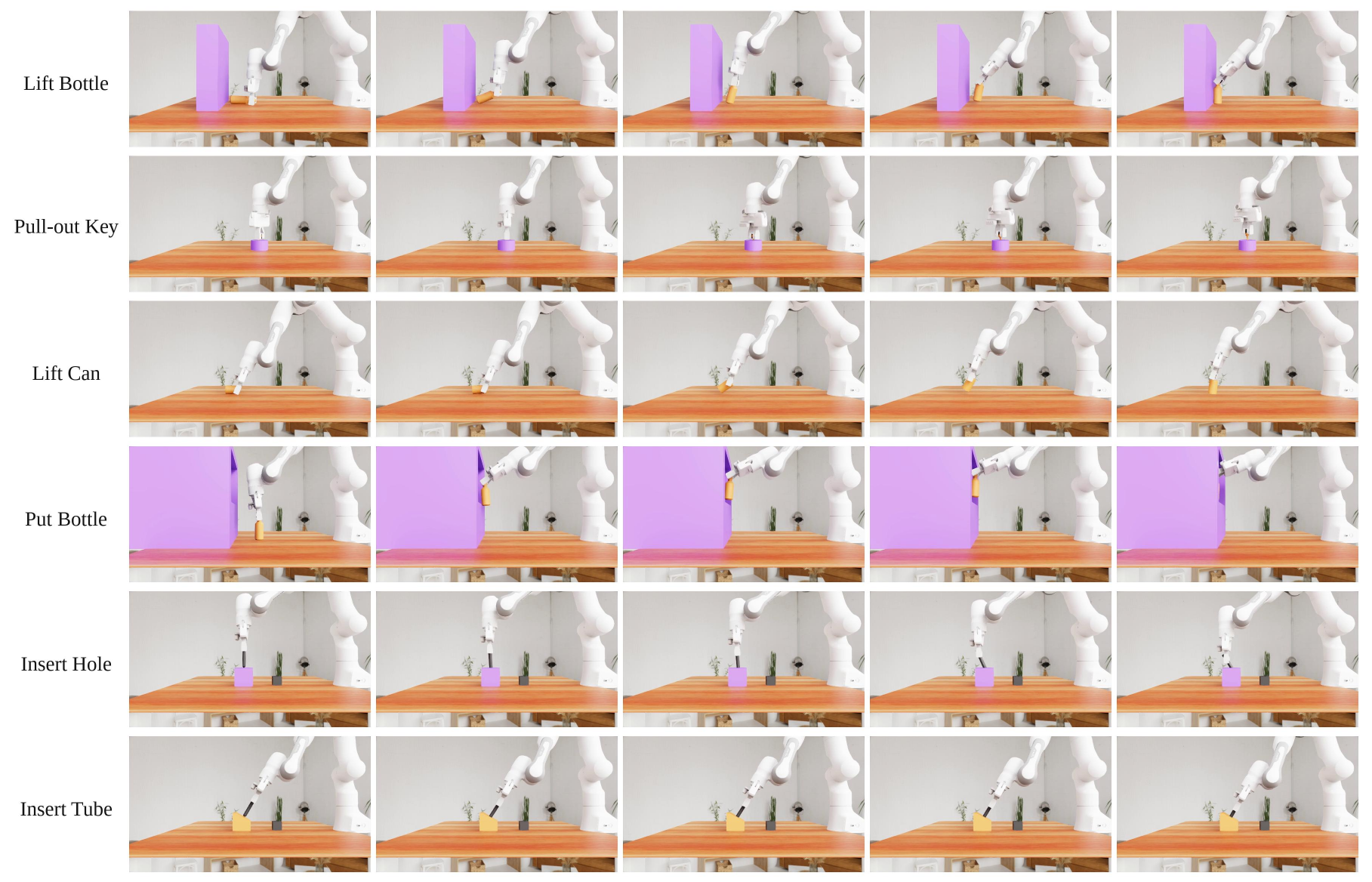}
  \vspace{-5mm}
  \caption{Representative rollouts on the UniVTAC simulation benchmark. Each row corresponds to one task and shows five key frames from a rollout.}
  \label{fig:appendix_rollout_univtac}
\end{figure*}

\begin{figure*}[t]
  \centering
  \includegraphics[width=\linewidth]{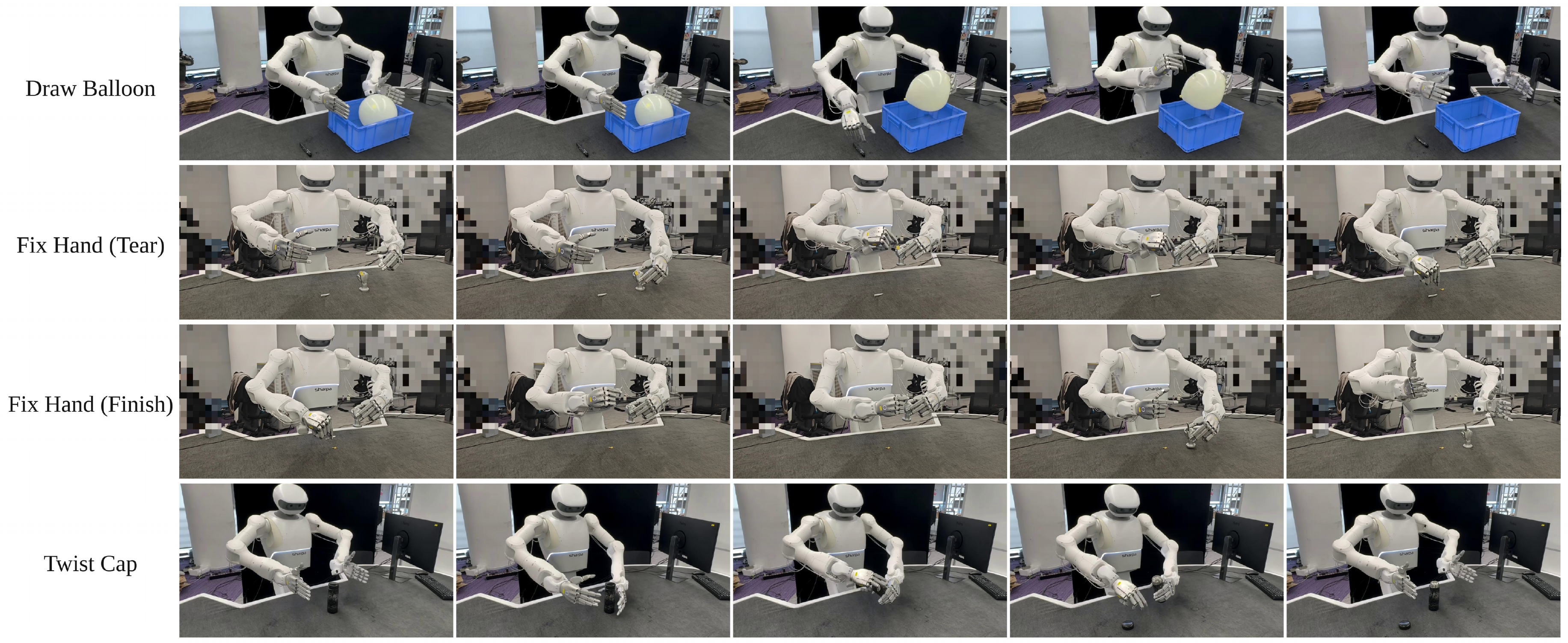}
  \vspace{-5mm}
  \caption{Representative rollouts on the Sharpa North setup, including Draw Balloon, Fix Hand (Tear), Fix Hand (Finish), and Twist Cap.}
  \label{fig:appendix_rollout_sharpanorth}
\end{figure*}

\begin{figure*}[t]
  \centering
  \includegraphics[width=\linewidth]{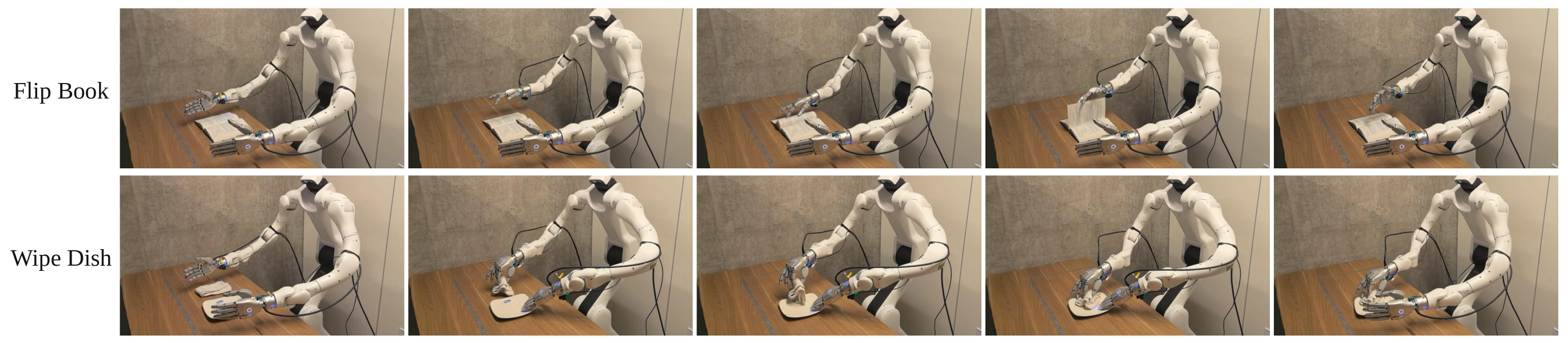}
  \vspace{-5mm}
  \caption{Representative rollouts on the Sharpa\&Dexmate setup, including Flip Book and Wipe Dish.}
  \label{fig:appendix_rollout_sharpadexmate}
\end{figure*}

We provide representative rollout visualizations for the seen-sensor evaluation setups in Fig.~\ref{fig:appendix_rollout_univtac}, Fig.~\ref{fig:appendix_rollout_sharpanorth}, and Fig.~\ref{fig:appendix_rollout_sharpadexmate}. Each row corresponds to one task and shows five key frames sampled from a successful rollout.

\subsection{Details of Similar Sensors for Unseen Sensor Evaluation}\label{app:similar_sensor}

\begin{figure*}[t]
  \centering
  \includegraphics[width=\linewidth]{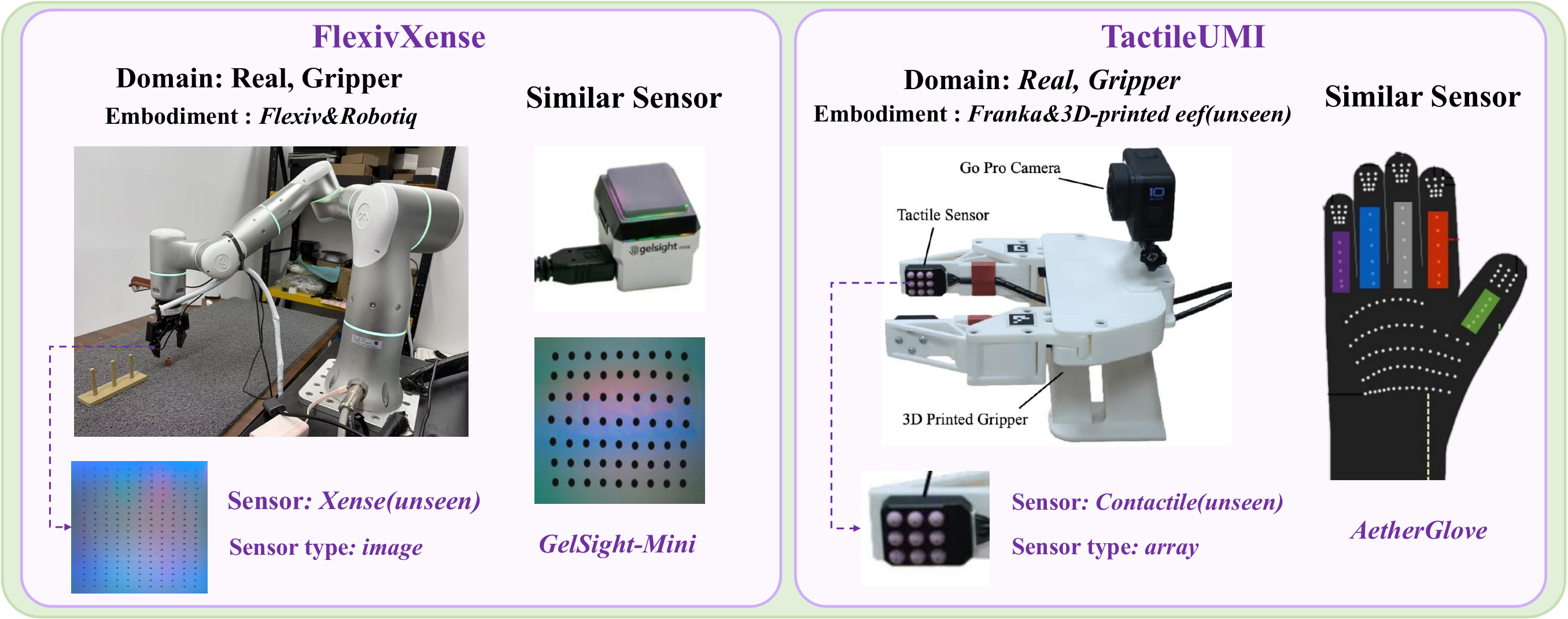}
  \vspace{-5mm}
  \caption{Similar sensors between the pretraining dataset and the downstream evaluation setups.}
  \label{fig:appendix_similar_sensors}
\end{figure*}

In our unseen sensor evaluation setup, Figure~\ref{fig:appendix_similar_sensors} shows the pretraining sensors that are most similar to the downstream evaluation sensors. For Xense (image-type), the closest counterpart is GelSight-Mini~\cite{yuan2017gelsight}. For Contactile~\cite{velasco2025touch} (array-type), the closest counterpart is AetherGlove.

\subsection{Visualization of Policy Rollouts for Unseen-Sensor Evaluation}
\label{app:unseen_sensor_rollout}

\begin{figure*}[t]
  \centering
  \includegraphics[width=\linewidth]{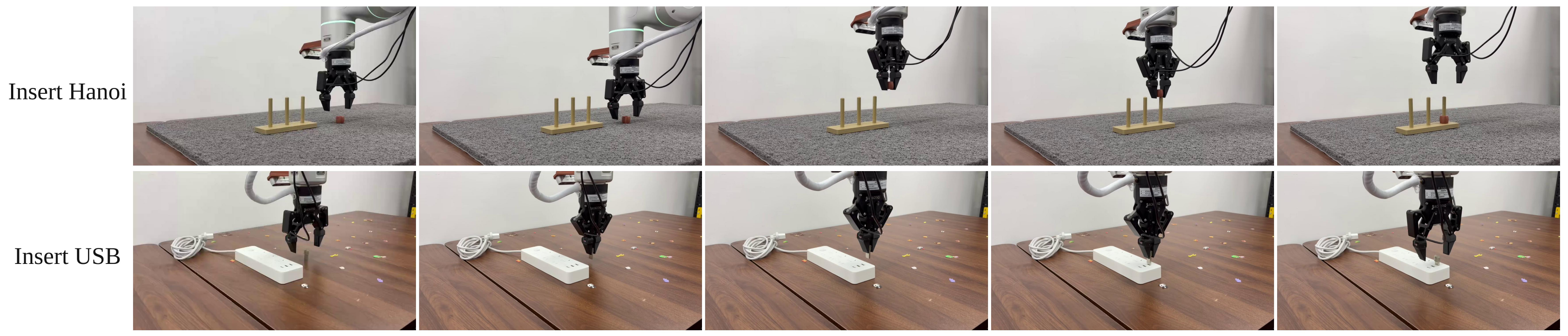}
  \vspace{-5mm}
  \caption{Representative rollouts on the FlexivXense setup, including Insert Hanoi and Insert USB.}
  \label{fig:appendix_rollout_flexivxense}
\end{figure*}

\begin{figure*}[t]
  \centering
  \includegraphics[width=\linewidth]{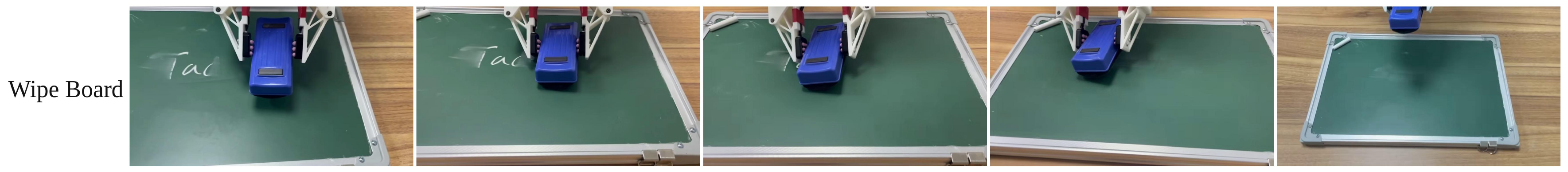}
  \vspace{-5mm}
  \caption{Representative rollouts on the TactileUMI setup for the Wipe Board task.}
  \label{fig:appendix_rollout_tactileumi}
\end{figure*}

We provide representative rollout visualizations for the unseen-sensor evaluation setups in Fig.~\ref{fig:appendix_rollout_flexivxense} and Fig.~\ref{fig:appendix_rollout_tactileumi}.

\subsection{Details of Pretrained Tactile Knowledge Ablation}\label{app:ablation_result}

Detailed ablation results for pretrained tactile knowledge are provided in Tab.~\ref{tab:univtac_results_ablation}. 

\begin{table}[t]
\centering
\caption{Results of Pretrained Tactile Knowledge Ablation on the UniVTAC simulation benchmark. We report success rates (\%). ``Avg. w/o Lifts'' excludes Lift Bottle and Lift Can. The best and second-best results are highlighted in \textbf{bold} and \underline{underline}, respectively.}
\vspace{-1mm}
\label{tab:univtac_results_ablation}
\setlength{\tabcolsep}{4pt}
\renewcommand{\arraystretch}{1.08}
\resizebox{\linewidth}{!}{
\begin{tabular}{lcccccccc}
\toprule
\textbf{Method}
& \textbf{Lift Bottle}
& \textbf{Pull-out Key}
& \textbf{Lift Can}
& \textbf{Put Bottle}
& \textbf{Insert Hole}
& \textbf{Insert Tube}
& \textbf{Avg.}
& \textbf{Avg. w/o Lift}\\
\midrule

\textbf{FTP-\boldmath$\pi_{0.5}$}
& 77 & 30 & 26 & 19 & \underline{47} & \underline{72} & 45.16 & \underline{42} \\

\textbf{NTP-1}
& \underline{88} & \underline{38} & \textbf{66} & \underline{32} & 31 & 45 & \underline{50.00} & 36.5 \\

\midrule
\textbf{FTP-1}
& \textbf{97} & \textbf{48} & \underline{65} & \textbf{47} & \textbf{64} & \textbf{79} & \textbf{66.66} & \textbf{59.5} \\
\bottomrule
\end{tabular}
}
\end{table}






\clearpage
\section{Additional Analyses}
\label{app:additional_analyses}

\subsection{Matched VLA Pretraining Baseline}
\label{app:matched_pretraining}

To examine the effect of exposure to the pretraining trajectories, we continue pretraining $\pi_{0.5}$ without tactile input on the same robot trajectories used by FTP-1, using the same pretraining budget. On the UniVTAC simulation benchmark~\cite{chen2026univtac}, this baseline achieves an average success rate of 57.0\%, compared with 66.7\% for FTP-1. Excluding Lift Bottle and Lift Can, the corresponding success rates are 44.0\% and 59.5\%. The improvements of 9.7 and 15.5 percentage points under matched trajectory exposure and pretraining budget support the value of learning from tactile inputs in FTP-1. The average excluding the two lift tasks is informative because these tasks are largely solvable using vision alone, as shown in Tab.~\ref{tab:univtac_results}.

\subsection{Tactile Pretraining Data and Training Steps}
\label{app:budget_scaling}

We evaluate downstream performance while jointly varying the amount of tactile pretraining data and the number of training steps. The 50\% setting uses half of the full pretraining data and 25,000 training steps; the 100\% setting uses the full data and 50,000 steps. The 0\% setting is FTP-$\pi_{0.5}$, which uses the FTP-1 architecture without FTP-1 pretraining. Table~\ref{tab:pretraining_budget} reports the average success rate on UniVTAC and the average excluding Lift Bottle and Lift Can.

\begin{table}[htbp]
\centering
\caption{Joint scaling of pretraining data and training steps on UniVTAC. Success rates are reported in percent. ``Avg. w/o Lift'' excludes Lift Bottle and Lift Can.}
\label{tab:pretraining_budget}
\begin{tabular}{lcccc}
\toprule
Setting & Data fraction & Training steps & Avg. & Avg. w/o Lift \\
\midrule
0\% (FTP-$\pi_{0.5}$) & 0\% & 0 & 45.2 & 42.0 \\
50\% & 50\% & 25,000 & 59.0 & 46.0 \\
100\% (FTP-1) & 100\% & 50,000 & \textbf{66.7} & \textbf{59.5} \\
\bottomrule
\end{tabular}
\end{table}

Increasing the setting from 0\% to 50\% improves the overall average by 13.8 percentage points and the average excluding lift tasks by 4.0 points. Increasing both data and training steps from 50\% to 100\% yields a further 7.7 and 13.5 points, respectively. Full pretraining therefore improves the two metrics by 21.5 and 17.5 points over FTP-$\pi_{0.5}$. Performance improves consistently with this joint increase in data and optimization steps, with larger gains on the non-lift tasks when moving from half to full pretraining.

\subsection{Tactile Feature Probing Evaluation}
\label{app:frozen_probes}

We evaluate frozen tactile features on force estimation and slip detection to examine the information available in the learned representations. Table~\ref{tab:frozen_feature_probes} compares T3~\cite{zhao2024transferable} with the FTP-1 trunk and tokenizer features. Force estimation uses linear probes and is measured by mean absolute error (MAE) in newtons; slip detection uses an attentive probe and is measured by F1. Results are averaged over probe-training seeds 42, 43, and 44.

\paragraph{Feature extraction.}
We use the GelSight Mini image encoder and shared image Transformer from the original T3 checkpoint or the FTP-1 checkpoint after 50,000 pretraining steps. The T3 and FTP-1 trunk features are the 768-dimensional \texttt{[CLS]} outputs of the shared image Transformer. FTP-1 tokenizer features are the 512-dimensional outputs of its subsequent image projection, before the functional-area embeddings and shared morphology projection. All feature extractors remain frozen, and prediction heads are trained on cached features. TacBench samples contain the current frame and the frame five steps earlier; FeelAnyForce samples contain a single image. T3 and FTP-1 inputs are resized to $224\times224$, with channel normalization computed from the training split.

\paragraph{Benchmark splits.}
For TacBench T1~\cite{higuera2024sparsh}, we train on the sphere and sharp-indenter batches 1 and 2 and evaluate on the held-out flat-indenter batch 1; 10\% of the training trajectories are reserved for validation. For TacBench T2, we use a \emph{released-data reconstruction} from sphere batches 1--6, with trajectory-disjoint training, validation, and test sets and 25,035 test samples. The T2 comparison uses the same split for all representations. For FeelAnyForce~\cite{shahidzadeh2025feelanyforce}, we use the \emph{released primitive-indenter split}, retaining its CSV partitions of 96,520 training, 10,725 validation, and 2,864 test samples.

\paragraph{Probe training and metrics.}
The force probe is one affine map over the ordered frame features. The slip probe uses a 768-dimensional attentive pooler with 12 heads, followed by a slip classifier and an auxiliary force-change head. We train probes using Adam with a learning rate of $10^{-4}$ and batch size 256 for up to 100 epochs, selecting checkpoints by validation loss with an early-stopping patience of 15 epochs. Feature standardization and force-target standardization use training samples only. Force MAE averages absolute errors over the three force components and test samples. Slip F1 is computed after thresholding the predicted slip probability at 0.4 and applying a three-consecutive-positive filter within each trajectory.

For the T2 reference, we evaluate the frozen SPARSH-MAE encoder~\cite{higuera2024sparsh} on the same reconstruction using its native two-frame input and the attentive probe. For the FeelAnyForce reference, we take the DINOv2 backbone from the released model trained with force and depth supervision, freeze it, and train the same linear force probe. This reference therefore measures force information accessible after task-specific supervision.

\begin{table}[htbp]
\centering
\caption{Frozen-feature tactile evaluation, averaged over three probe-training seeds. T1 uses the held-out flat indenter, T2 uses the released-data reconstruction, and FeelAnyForce uses the released primitive-indenter split. The references are SPARSH-MAE for T2 and the force/depth-supervised FeelAnyForce backbone for force estimation.}
\label{tab:frozen_feature_probes}
\setlength{\tabcolsep}{5pt}
\small
\begin{tabular}{lcccc}
\toprule
Benchmark and metric & T3 & FTP-1 trunk & FTP-1 tokenizer & Reference \\
\midrule
T1 force: linear MAE $\downarrow$ (N) & 0.106 & 0.107 & 0.116 & --- \\
T2 slip: attentive F1 $\uparrow$ & 0.725 & 0.780 & 0.786 & 0.922 \\
FeelAnyForce: linear MAE $\downarrow$ (N) & 0.273 & 0.314 & 0.424 & 0.041 \\
\bottomrule
\end{tabular}
\end{table}

On T1 force estimation, the FTP-1 trunk achieves an MAE of 0.107\,N, close to the 0.106\,N of its T3 initialization, indicating that quantitative force information remains accessible after policy pretraining. On the T2 reconstruction, mean slip F1 increases from 0.725 for T3 to 0.780 for the FTP-1 trunk and 0.786 for the tokenizer. The attentive probe can therefore recover manipulation-relevant contact dynamics from the frozen FTP-1 features, complementing the downstream policy evaluations.

\end{document}